\RequirePackage{fix-cm}
\documentclass[smallextended]{svjour3}       
\smartqed  
\usepackage{graphicx}
\usepackage{amsmath}
\usepackage{amssymb}
\usepackage{graphicx}
\usepackage{epsfig}
\usepackage{nopageno}
\usepackage{color, verbatim}
\usepackage{enumerate}
\usepackage{fancyhdr}
\usepackage{ifthen}
\usepackage{stmaryrd}
\usepackage{algorithm2e}
\usepackage{url}

\begin{document}
\title{Contextual and Possibilistic Reasoning \\ for Coalition Formation}

\author{Antonis Bikakis         \and
        Patrice Caire 
}


\institute{A. Bikakis \at
              Department of Information Studies, UCL, UK \\
              \email{a.bikakis@ucl.ac.uk}           
           \and
           P. Caire \at
              SnT and CSC, University of Luxembourg, Luxembourg \\
              \email{patrice.caire@uni.lu}           
}

\date{Published: 19 June 2020}

\maketitle
\begin{abstract}
In multiagent systems, agents often have to rely on other agents to reach their goals, for example when they lack a needed resource or do not have the capability to perform a required action. Agents therefore need to cooperate. Then, some of the questions raised are: Which agent(s) to cooperate with? What are the potential coalitions in which agents can achieve their goals? As the number of possibilities is potentially quite large, how to automate the process? And then, how to select the most appropriate coalition, taking into account the uncertainty in the agents' abilities to carry out certain tasks? In this article, we address the question of how to find and evaluate coalitions among agents in multiagent systems using MCS tools, while taking into consideration the uncertainty around the agents' actions. Our methodology is the following: We first compute the solution space for the formation of coalitions using a contextual reasoning approach. Second, we model agents as contexts in Multi-Context Systems (MCS), and dependence relations among agents seeking to achieve their goals, as bridge rules. Third, we systematically compute all potential coalitions using algorithms for MCS equilibria, and given a set of functional and non-functional requirements, we propose ways to select the best solutions. Finally, in order to handle the uncertainty in the agents' actions, we extend our approach with features of possibilistic reasoning. We illustrate our approach with an example from robotics.

\end{abstract}

\section{Introduction}

In multiagent systems, agents have goals to satisfy. Typically, agents cannot reach all their goals by themselves, without any help. Instead, agents need to cooperate with other agents, for example because they need a specific resource to satisfy a goal, or do not have the capability required to perform a task. Moreover, in most real-world settings, we cannot always be certain that agents will carry out their assigned tasks successfully.

The questions then, are: Which agent to cooperate with? Which group of agents to join? The problem of assembling a group of cooperating agents in order for all agents to reach their goals, shared or not, is referred to as coalition formation, and has been on the focus of many recent studies in the area of multiagent systems 
(e.g., \cite{DBLP:journals/jasss/Sichman98,DBLP:conf/atal/SichmanC02,DBLP:journals/ai/ShehoryK98,klusch02,DBLP:journals/igpl/BoellaST09,DBLP:conf/atal/GrossiT10,DBLP:conf/atal/CaireVBT08}). This paper introduces a novel contextual reasoning approach to address the problem based on the use of Multi-Context Systems (MCS). 

\emph{Multi-Context Systems} (\emph{MCS})  \cite{giunchiglia94,ghidini01,brewkaEiter07} are logical formalizations of distributed context theories connected through a set of bridge rules, which enable information flow between different contexts. A \emph{context} can be thought of as a logical theory - a set of axioms and inference rules - that models the knowledge of an agent. Intuitively, MCS can be used to represent any information system that consists of heterogeneous knowledge-based agents including peer-to-peer systems, distributed ontologies or Ambient Intelligence systems. Several applications have already been developed on top of MCS or other similar formal models of context including (\emph{a}) the CYC common sense knowledge base \cite{lenat89}, (\emph{b}) contextualized ontology languages, such as Distributed Description Logics \cite{borgida03} and C-OWL \cite{bouquet03}, (\emph{c}) context-based agent architectures \cite{parsons98,sabater02}, and (\emph{d}) distributed reasoning algorithms for Mobile Social Networks \cite{antoniouKR10} and Ambient Intelligence systems \cite{bikakisKAIS11}.

In this article, we address the question of how to find and evaluate coalitions among agents while taking advantage of the MCS model and algorithms. Specifically, our approach uses two variants of MCS. The first variant, called \emph{nonmonotonic MCS} \cite{brewkaEiter07}, allows us to handle incomplete information and potential conflicts that may arise when integrating information from different sources. The second variant, called \emph{possibilistic MCS} \cite{yin12}, is a formalism that enables to model uncertainty in context theories and bridge rules. The main advantages of our approach are: (a) MCS can represent heterogenous multiagent systems, i.e. systems containing agents with different knowledge representation models; (b) bridge rules can represent different kinds of inter-agent relationships such as dependencies, constraints and conflicting goals; (c) the possibilistic extension of MCS enables modeling uncertainty in the agents' actions; (d) there are both centralized and distributed algorithms that can be used for computing the potential coalitions. 

We formulate our main research question as: 
\begin{itemize}
\item \emph{How to find and evaluate coalitions among agents in multiagent systems using MCS tools while taking into consideration the uncertainty around the agents' actions?}
\end{itemize}

This breaks down into the following three sub-questions:
\begin{enumerate} 
\item \emph{How to formally compute the solution space for coalition formation using the MCS model and algorithms?}
\item \emph{How to select the best solution given a set of requirements?}
\item \emph{How to compute and evaluate coalitions taking also into account the uncertainty in the agents' actions?}
\end{enumerate}

Our methodology is the following. We start with modeling dependencies among agents using dependence relations as described in \cite{DBLP:conf/atal/SichmanC02}. We then model the system as a nonmonotonic MCS: each agent is modeled as a context with a knowledge base and an underlying logic and dependence relations are modeled as bridge rules. Third, we use appropriate algorithms to compute MCS equilibria. Each equilibrium corresponds to a different coalition. Finally, given a set of requirements, we show how to select the best solutions. The requirements we consider may be of two kinds. They may be domain related. For example in robotics, power consumption is a key concern that must be carefully dealt with. They may also be system related. For example in multiagent systems, the efficiency and conviviality of the system may be considered. We then extend our approach with features of possibilistic reasoning: We extend the definition of dependence relations with a certainty degree; we then use the model and algorithms of possibilistic MCS to compute the potential coalitions under uncertainty. In this case, we evaluate the different coalitions based on the certainty degree with which each coalition achieves the different goals, using multiple criteria decision making methods. 

This article is an extended version of \cite{eumas14}, where we presented our methodology for formalizing multiagent systems and computing coalitions between agents under the perfect world assumption, i.e. actions are always carried out with success by the agents that they have been assigned to. Here, we provide more details about the computation of coalitions and the selection of the best coalition in the perfect world case. We also present new results for the cases that the perfect world assumption does not hold due to uncertainty around the agents' actions.

The rest of the paper is structured as follows. Section 2 presents background information on dependence networks, coalition formation, nonmonotonic MCS and possibilistic MCS using an example from social networks. Section 3 introduces our main example originated in robotics. Section 4 describes our approach in a setting without uncertainty: how we use MCS to represent agents and their dependencies; how we systematically compute the coalitions; and how we then select the best coalitions with respect to given requirements. Section 5 presents the possibilistic reasoning approach, which takes into account the uncertainty in the agents' actions. Section 6 presents related research, and Section 7 concludes with a summary and a perspective on future works.

\section{Background} \label{sec:background}

\subsection{Dependence Networks and Coalition Formation}

Our model for dependencies among agents in multiagent systems is based on dependence networks. According to Conte and Sichman \cite{Sichman01onsocial}, dependence networks can be used to represent the pattern of relationships that exist between agents, and more specifically, interdependencies among agents' goals and actions. They can be used to study emerging social structures such as aggregates of heterogeneous agents. They are based on a social reasoning mechanism, on social dependence and on power \cite{DBLP:conf/atal/SichmanC02}. Power, in this context, means the ability to fulfill a goal. Multi-agent dependence allows one to express a wide range of interdependent situations between agents. 

A dependence network consists of a finite set or sets of actors and the relation or relations between
them \cite{DBLP:conf/ecai/SichmanCCD94}. Actors can be people or organizations. They are linked together by goals, behaviors and exchanges such as hard currency or information. The structural similarity between dependence networks and directed graphs is such that a dependence network can be represented as a directed graph. Informally, the nodes in the graph represent both the agents themselves, and the actions they have to perform to reach a goal. The directed edges in the graph are labelled with goals, and link agents with actions.

When agents cooperate to achieve some of their goals, they form groups or coalitions. Coalitions are topological aspects of a dependence network. They are indicative of some kind of organization, for example, the cooperation between agents in the dependence network. The coalition is supposed to ensure individual agents a sufficient payoff to motivate them to collaborate. In a coalition, agents coordinate their behaviors to reach their shared or reciprocal goals, as for example described in \cite{sauro:thesis,Sichman01onsocial}. All the agents in the coalition somehow benefit from the goals being reached. A coalition can achieve its purpose if its members are cooperative, i.e., if they adopt the goals of the coalition in addition to their own goals.

\subsection{Multi-Context Systems}

Multi-Context Systems (MCS)  \cite{giunchiglia94,ghidini01,brewkaEiter07} has been the main effort to formalize context and contextual reasoning in Artificial Intelligence. We use here the definition of heterogeneous nonmonotonic MCS given in \cite{brewkaEiter07}. The main idea is to allow different logics to be used in different contexts, and to model information flow among contexts via bridge rules. 

\subsubsection{Formalization}

According to \cite{brewkaEiter07}, a MCS is a set of contexts, each composed of a knowledge base with an underlying logic, and a set of bridge rules. A logic \emph{L} = (\textbf{KB}$_L$, \textbf{BS}$_L$, \textbf{ACC}$_L$) consists of the following components:

\begin{itemize}
\item \textbf{KB}$_L$ is the set of well-formed knowledge bases of \emph{L}. Each element of \textbf{KB}$_L$ is a set of formulae.
\item \textbf{BS}$_L$ is the set of possible belief sets, where the elements of a belief set is a set of formulae. 
\item \textbf{ACC}$_L$:  \textbf{KB}$_L$ $\rightarrow$ $2^{\textrm{BS}_L}$ is a function describing the semantics of the logic by assigning to each knowledge base a set of acceptable belief sets.
\end{itemize}
As shown in \cite{brewkaEiter07}, this definition captures the semantics of many different logics both monotonic, e.g. propositional logic, description logics and modal logics, and nonmonotonic, e.g. default logic, circumscription, defeasible logic and logic programs under the answer set semantics. 

A \emph{bridge rule} refers in its body to other contexts and can thus add information to a context based on what is believed or disbelieved in other contexts. Bridge rules are added to those contexts to which they potentially add new information. Let $\mathcal{L}$ = (\emph{L}$_1$, $\ldots$, \emph{L}$_n$) be a sequence of logics. An $L_k$-bridge rule $r$ over $\mathcal{L}$, $1\leq k \leq n$, is of the form
\begin{equation}\label{eq:bridge}
\begin{split}
r  = & (k:s)\leftarrow (c_1:p_1), \ldots, (c_j:p_j), \\
& \textbf{not} (c_{j+1}:p_{j+1}), \ldots , \textbf{not} (c_m:p_m).
\end{split}
\end{equation}

\noindent where $c_i$, $\textrm{1}\leq i \leq n$, refers to a context, $p_i$ is an element of some belief set of $L_i$, and $k$ refers to the context receiving information $s$. We denote by $h_b(r)$ the belief formula $s$ in the head of $r$.

A \emph{MCS} $M=(c_1, \ldots , c_n)$ is a set of contexts $c_i = (L_i,kb_i,br_i)$,  $\textrm{1} \leq i \leq n$, where \emph{L$_i$} = (\textbf{KB}$_i$, \textbf{BS}$_i$, \textbf{ACC}$_i$) is a logic, $kb_i \in \textbf{KB}_i$ a knowledge base, and $br_i$ a set of $L_i$-bridge rules over (\emph{L}$_1$, $\ldots$, \emph{L}$_n$). For each $H \subseteq \{h_b(r) \mid r \in br_i \}$ it holds that $kb_i\cup H \in \textbf{KB}_{i}$, meaning that bridge rule heads are compatible with knowledge bases.

A belief state of a MCS is the set of the belief sets of its contexts. Formally, a \emph{belief state} of $M=(c_1, \ldots , c_n)$ is a sequence $\mathcal{S}=(S_1, \ldots , S_n)$ such that $S_i \in \textbf{BS}_i$. Intuitively, $\mathcal{S}$ is derived from the knowledge of each context and the information conveyed through applicable bridge rules. A bridge rule of form (\ref{eq:bridge}) is applicable in a belief state $\mathcal{S}$ iff for $\textrm{1}\leq i \leq j$: $p_i \in S_{i}$ and for $j < l \leq m$: $p_l \notin S_{l}$. 

Equilibrium semantics selects certain belief states of a MCS as acceptable. Intuitively, an equilibrium is a belief state $\mathcal{S}=(S_1, \ldots , S_n)$ where each context $c_i$ respects all bridge rules applicable in $\mathcal{S}$ and accepts $S_i$. Formally,  $\mathcal{S}=(S_1, \ldots , S_n)$ is an equilibrium of $M$, iff for $\textrm{1}\leq i \leq n$, $$S_i \in \textbf{ACC}_i(kb_i \cup \{h_b(r) \mid r\in br_i \textrm{ applicable in } \mathcal{S} \} ).$$
$\mathcal{S}$ is a grounded equilibrium of $M$ iff for $\textrm{1}\leq i \leq n$, $S_i$ is an answer set of logic program $P=kb_i \cup \{h_b(r) \mid r\in br_i \textrm{ applicable in } \mathcal{S} \}$. For a definite MCS (MCS without default negation in the bridge rules), its unique grounded equilibrium is the collection consisting of the least (with respect to set inclusion) Herbrand model of each context.

In a MCS, even if  contexts are locally consistent, their bridge rules may render the whole system inconsistent. This is formally described in \cite{brewkaEiter07} as a \emph{lack of an equilibrium}. Most methods for inconsistency resolution in MCS are based on the following intuition: a subset of the bridge rules that cause inconsistency must be invalidated and another subset must be unconditionally applied, so that the entire system becomes consistent again (e.g. \cite{eiterKR10,eiterJELIA10}). In \cite{prima13} we proposed a different method based on \emph{conviviality}, a property of multiagent systems.

\begin{example} \label{MCSexample}
Consider a scholar social network through which software agents, acting on behalf of students or researchers, share information about research articles they find online. Consider three such agents, each one with its own knowledge base and logic exchanging information about a certain article. The three agents can be represented as contexts $c_1-c_3$ in a MCS $M=\{c_1,c_2,c_3\}$. The knowledge bases of the three contexts are respectively:
\begin{equation*}
\begin{split} 
kb_1 = & \left \{ \begin{array}{l} sensors \leftarrow, \: corba \leftarrow, \\  distributedComputing \leftarrow corba, not\: centralizedComputing \end{array} \right \} \\
kb_2 =  & \{  profA \} \\
kb_3 = & \{ ubiquitousComputing \subseteq ambientComputing \}
\end{split} 
\end{equation*}

\noindent $kb_1$ is a logic program stating that the article is about \emph{sensors} and \emph{corba}, and that articles about \emph{corba} that are not classified in \emph{centralizedComputing} can be classified in \emph{distributedComputing}. $kb_2$ states in propositional logic that the article is written by \emph{profA}. $kb_3$ is an ontology about computing written in a basic description logic, according to which \emph{ubiquitousComputing} is a type of \emph{ambientComputing}. The three agents use bridge rules $r_1$-$r_4$ to exchange information about articles.

\begin{equation*}
\begin{split} 
r_1 = & (c_1 : centralizedComputing) \leftarrow  (c_2 : middleware)\\
r_2 = & (c_1 : distributedComputing) \leftarrow  (c_3 :ambientComputing) \\
r_3 = & (c_2 : middleware) \leftarrow (c_1 : corba)\\
r_4 = &  (c_3: ubiquitousComputing) \leftarrow   (c_1 : sensors), (c_2 : profB)
\end{split} 
\end{equation*}

\noindent With $r_1$ and $r_2$, the first agent classifies articles about \emph{middleware} (as described in $c_2$) in the category of \emph{centralizedComputing}, and articles about \emph{ambientComputing} (as described in $c_3$) in \emph{distributedComputing}. With $r_3$, the second agent classifies articles about \emph{corba} in  \emph{middleware}. Finally, with $r_4$, the third agent classifies articles about \emph{sensors}, which have been written by \emph{profB}, in \emph{ubiquitousComputing}. $M$ has one equilibrium:
\begin{equation*}
\begin{split} 
\mathcal{S} = & \left ( \begin{array}{l} \{sensors,corba,centralizedComputing\},\\ \{profA,middleware\}, \emptyset \end{array} \right )
\end{split} 
\end{equation*}

\noindent according to which, the first agent classifies the paper in \emph{centralizedComputing}, and the second agent classifies it in \emph{middleware}.

Consider now the case that \emph{profB} is identified by $c_2$ as a second author of the paper: $$kb_2=\{profA,profB\}$$
Rules $r_4$ and $r_2$ would then become applicable, and as a result $M$ would not have an equilibrium; it would therefore be inconsistent. To resolve the conflict, one of the four bridge rules $r_1$-$r_4$ would have to be invalidated. For example, by invalidating rule $r_1$, the system would have one equilibrium:

\begin{equation*}
\begin{split} 
\mathcal{S}_1 =  & \left ( \begin{array}{l}\{sensors,corba, distributedComputing\},\\  \{profA,profB,middleware\}, \\ \{ubiquitousComputing,ambientComputing\} \end{array} \right )
\end{split} 
\end{equation*}
\end{example}

\subsubsection{Computational Complexity}

Paper \cite{brewkaEiter07} presents an analysis on computational complexity, focusing on MCS with logics that have \emph{poly-size kernels}. A logic $L$ has poly-size kernels, if there
is a mapping $\kappa$, which assigns to every $kb \in$ \textbf{KB} and $S \in$ \textbf{ACC}$(kb)$ a set $\kappa (kb,S) \subseteq S$ of size (written as a string) polynomial in the size of $kb$, called the \emph{kernel of} $S$, such that there is a one-to-one correspondence $f$ between the belief sets in \textbf{ACC}$(kb)$ and their kernels, i.e., $S \rightleftharpoons f(\kappa(kb, S))$. Examples of logics with poly-size kernels include propositional logic, default logic, autoepistemic logic and nonmonotonic logic programs. If furthermore, given any knowledge base $kb$, an element $b$, and a set of elements $K$, deciding whether (i) $K = \kappa(kb,S)$ for some $S \in$ \textbf{ACC}$(kb)$ and (ii) $b \in S$ is in $\Delta _{k}^{p}$, then we say that $L$ has kernel reasoning in $\Delta _{k}^{p}$. For example, default logic and autoepistemic logic have kernel reasoning in $\Delta _{2}^{p}$. 

According to the analysis in \cite{brewkaEiter07}, for a finite MCS $M$ (where all knowledge bases $kb_i$ and sets of bridge rules $br_i$ are finite, and the logics $L_i$ are from an arbitrary but fixed set), where all logics $L_i$ have poly-size kernels and kernel reasoning in $\Delta _{k}^{p}$, deciding whether a literal $p$ is in a belief set $S_i$ for some (or each) equilibrium of $M$ is in $\Sigma _{k+1}^{p}$  (resp. $\Pi _{k+1}^{p}=co-\Sigma _{k+1}^{p}$).

\subsection{Possibilistic reasoning in MCS}

Recently, Yin et al. proposed a framework for possibilistic reasoning in Multi-Context Systems, which they called \emph{possibilistic MCS} \cite{yin12}. This has been so far the only attempt to model uncertainty in MCS. It is based on \emph{possibilistic logic} \cite{dubois94} and \emph{possibilistic logic programs} \cite{nicolas06}, which are logic-based frameworks for representing states of partial ignorance using a dual pair of possibility and necessity measures. These frameworks are in turn based on ideas from Zadeh's possibility theory \cite{zadeh78}. Below, we first provide some preliminary information on possibilistic logic programs, which will then help us to present possibilistic MCS.

\subsubsection{Possibilistic Logic Programs \cite{nicolas06}} 

Possibilistic logic and logic programs use the notion of \emph{possibilistic concept}, which is denoted by $\overline{X}$, where $X$ denotes its classical counterpart. For example, in possibilistic logic programs, this notion is used in the definitions of \emph{possibilistic atoms} and \emph{poss-programs}:

\begin{definition} \label{def:PossAtom}
Let $\Sigma$ be a finite set of atoms. A possibilistic atom is $\overline p = (p,[\alpha])$, where $p \in \Sigma$ and $\alpha \in [0,1]$.
\end{definition}

The classical projection of $\overline p$ is the atom $p$ and $n(p)=\alpha$ is called the necessity degree of $\overline p$.

\begin{definition} \label{def:PossProg}
A possibilistic normal logic program (or poss-program) $\overline p$ is a set of possibilistic rules of the form:
\begin{equation}\label{eq:possRule}
\overline r  = s \leftarrow p_1, \ldots, p_m, not \; q_1, \ldots , not \; q_n, [\alpha].
\end{equation}
where $m,n \geq 0$, $\{p_1, \ldots, p_m, q_1, \ldots , q_n, s\} \subseteq \Sigma$, and $n(\overline r)=\alpha \in [0,1]$.
\end{definition}

In (\ref{eq:possRule}), $\alpha$ represents the certainty level of the information described by rule $\overline r$. The head of  $\overline r$ is defined as $head(\overline r)=s$ and its body as $body(\overline r)=body^+(\overline r) \cup not \; body^-(\overline r)$, where $body^+(\overline r)=\{p_1, \ldots, p_m\}$ and $body^-(\overline r)=\{q_1, \ldots, q_n\}$. The positive projection of $\overline r$ is 
\begin{equation}
\overline r^+=head(\overline r) \leftarrow body^+(\overline r), [\alpha]
\end{equation}
The classical projection of $\overline r$ is the classical rule: 
\begin{equation}
r  = s \leftarrow p_1, \ldots, p_m, not \; q_1, \ldots , not \; q_n
\end{equation}

If a poss-program $\overline p$ does not contain any default negation then $\overline p$ is called a \emph{definite poss-program}. The reduct of a poss-program $\overline p$ w.r.t. a set of atoms $T$ is the definite poss-program defined as:
\begin{equation}\label{ruleReduct}
\overline p^{T} = \{\overline r^+ \mid \overline r \in \overline p, body^-(\overline r) \cap T = \emptyset \}
\end{equation}

For a set of atoms $T\subseteq \Sigma$ and a rule $\overline r \in \overline p$, we say that $\overline r$ is applicable in $T$ if $body^+(\overline r) \subseteq T$ and $body^-(\overline r) \cap T = \emptyset$. $App(\overline p,T)$ denotes the set of rules in $\overline p$ that are applicable in $T$. 

$P$ is said to be \emph{grounded} if it can be ordered as a sequence $\langle \overline r_1, \ldots , \overline r_n \rangle$ such that 
\begin{equation}
\forall i, 1 \geq i \geq n, \overline r_i \in App(\overline p,head(\{\overline r_1, \ldots , \overline r_{i-1}\}))
\end{equation}

Given a poss-program $\overline p$ over a set of atoms $\Sigma$, the semantics of $\overline p$ is defined through possibility distributions on $\Sigma$.\footnote{For more details about the semantics of poss-programs, see \cite{nicolas06}.}

\subsubsection{Possibilistic MCS \cite{yin12}}

A \emph{possibilistic MCS} (or \emph{poss-MCS}) is a collection of possibilistic contexts. A \emph{possibilistic context} $\overline c$ is a triple $(\Sigma,\overline P, \overline B)$ where $\Sigma$ is a set of atoms, $\overline P$ is a poss-program, and $\overline B$ is a set of possibilistic bridge rules. A \emph{possibilistic bridge rule} is defined as follows:

\begin{definition}\label{def:PossBridge}

Let $\overline k, \overline{c_1}, \dots ,\overline{c_n}$ be possibilistic contexts. A possibilistic bridge rule $\overline{pr}$ for context $\overline k$ is of the form
\begin{equation}\label{eq:possBridge}
\begin{split}
\overline{pr}  = & (\overline k:s)\leftarrow (\overline{c_1}:p_1), \ldots, (\overline{c_j}:p_j), \\
& \textbf{not} (\overline c_{j+1}:p_{j+1}), \ldots , \textbf{not} (\overline{c_m}:p_m), [\alpha]
\end{split}
\end{equation}
where $s$ is an atom in $\overline k$ and each $p_i$ is an atom in context $\overline{c_i}$. $1\leq i \leq n$. 
\end{definition}

Intuitively, a rule of form (\ref{eq:possBridge}) states that information $s$ is added to context $\overline k$ with necessity degree $\alpha$ if, for $1\geq i \geq j$, $p_i$ is provable in context $\overline{c_i}$ and for $j+1 \geq l \geq n$, $p_l$ is not provable in $\overline{c_l}$.

By $pr$ (see equation (\ref{eq:bridge})) we denote the classical projection of $\overline{pr}$. The necessity degree of $\overline{pr}$ is denoted by $n(\overline{pr})$.

\begin{definition}\label{def:PossMCS}

A possibilistic Multi-Context System, or just poss-MCS, $\overline M = (\overline{c_1},\ldots,\overline{c_n})$ is a collection of possibilistic contexts $\overline{c_i} = (\Sigma_i,\overline{P_i}, \overline{B_i})$,  $1\geq i \geq n$, where each $\Sigma_i$ is the set of atoms used in context $\overline{c_i}$, $\overline{P_i}$ is a poss-program on $\Sigma_i$ and $\overline{B_i}$ is a set of possibilistic bridge rules over atom sets $(\Sigma_i,\ldots, \Sigma_n)$.

A poss-MCS is definite if the poss-program and possibilistic bridge rules of each context are definite.

\end{definition}

\begin{definition}\label{def:PossBS}

A possibilistic belief state, $\overline{\mathcal{S}} = (\overline{S_1},\ldots,\overline{S_n})$ is a collection of possibilistic atom sets $\overline{S_i}$, where each $\overline{S_i}$ is a collection of possibilistic atoms $\overline{p_i}$ and $p_i \in \Sigma_i$.

\end{definition}

We will now describe the semantics for poss-MCS, starting with definite poss-MCS. The following definition specifies the possibility distribution of belief states for a given definite poss-MCS. It uses the notion of \emph{satisfiability} of a rule $r$, which is based on its applicability w.r.t. a belief state $\mathcal{S}$: 
\begin{equation}
\mathcal{S}\not\models r \mbox{ iff }body^+(r)\subseteq \mathcal{S} \mbox{ and } head(r) \not\in \mathcal{S}
\end{equation}

\begin{definition}\label{def:PossDistribution}

Let $\overline M = (\overline{c_1},\ldots,\overline{c_n})$ be a definite poss-MCS and $\overline{\mathcal{S}} = (\overline{S_1},\ldots,\overline{S_n})$ a belief state. The possibility distribution $\pi_{\overline M}: 2^\Sigma \rightarrow [0,1]$ for $\overline M$ is defined as:
\begin{equation}\label{eq:possDistribution}
\begin{split}
\pi_{\overline M}(\mathcal{S}) = \begin{cases} 0, \mbox{     if } \mathcal{S} \not \subseteq head(\bigcup \limits_i App_i(M,\mathcal{S})) \\ 0,  \mbox{     if } \bigcup \limits_i App_i(M,\mathcal{S}) \mbox{ is not grounded} \\ 1, \mbox{     if } \mathcal{S} \mbox{ is an equilibrium of } M \\ 1-max\{n(\overline r) \mid \mathcal{S} \not\models \overline r, \overline r \in \overline B_i \cup \overline L_i\}, \mbox{ otherwise} \end{cases}
\end{split}
\end{equation}
\end{definition}

The possibility distribution specifies the degree of compatibility of each belief set $S$ with the poss-MCS $\overline M$. Based on definition \ref{def:PossDistribution} we can now define the possibility and necessity of an atom is a belief state $S$.

\begin{definition}\label{def:AtomPossNec}

Let $\overline M$ be a definite poss-MCS and $\pi_{\overline M}$ be the possibilistic distribution for $\overline M$. The possibility and necessity of an atom $p_i$ in a belief state $\mathcal{S}$ are respectively defined as:
\begin{equation}\label{eq:atomPossibility}
\begin{split}
\Pi_{\overline M}(p_i) = max \{\pi_{\overline M}(\mathcal{S})\mid p_i \in S_i\}
\end{split}
\end{equation}
\begin{equation}\label{eq:atomNecessity}
\begin{split}
N_{\overline M}(p_i) = 1-max \{\pi_{\overline M}(\mathcal{S})\mid p_i \not\in S_i\}
\end{split}
\end{equation}
\end{definition}

$\Pi_{\overline M}(p_i)$ represents the level of consistency of $p_i$ w.r.t. the poss-MCS $\overline M$, while $N_{\overline M}(p_i)$ represents the level at which $p_i$ can be inferred from $M$. For example, whenever an atom $p_i$ belongs to the equilibrium of $M$ (the classical projection of $\overline M$), its possibility is equal to 1.

The semantics for definite poss-MCS is determined by its unique possibilistic grounded equilibrium.

\begin{definition}\label{def:PossGrEq}

Let $\overline M$ be a definite poss-MCS. Then the following set of possibilistic atoms is referred to as the possibilistic grounded equilibrium: 
\begin{equation}
\overline{MD} (\overline M) = \{\overline{S_1},\ldots, \overline{S_n}\}
\end{equation}
where $\overline{S_i}=\{(p_i,N_{\overline M}(p_i))\mid p_i \in \Sigma_i, N_{\overline M}(p_i)>0\}$ for $i=1,\ldots,n.$
\end{definition}

As proved in \cite{yin12} (Proposition 5), the classical projection of $\overline M \overline D (\overline M)$ is the grounded equilibrium of $M$, where $M$ is the classical projection of $\overline M$.

The definition of the semantics for normal poss-MCS is based on the notion of reduct for normal poss-MCS, which is in turn based on the definition of rule reduct (see equation (\ref{ruleReduct})):

\begin{definition}\label{def:PossMCSReduct}

Let $\overline M = (\overline{c_1},\ldots,\overline{c_n})$ be a normal poss-MCS and $\mathcal{S} = (S_1,\ldots,S_n)$ a belief state. The possibilistic reduct of $\overline M$ w.r.t. $\mathcal{S}$ is the poss-MCS
\begin{equation}
\overline M^{\mathcal{S}} = (\overline{c_1}^{\mathcal{S}},\ldots,\overline{c_n}^{\mathcal{S}})
\end{equation}
where $\overline{c_i}^{\mathcal{S}} = (\Sigma_i, \overline{P_i}^{\mathcal{S}}, \overline{B_i}^{\mathcal{S}})$.
\end{definition}

Note that the reduct of $\overline{P_i}$ relies only on $S_i$ while the reduct of $\overline{B_i}$ depends on the whole belief state $\mathcal{S}$.

Given the notion of reduct for normal poss-MCS, the equilibrium semantics of normal poss-MCS is defined as follows:

\begin{definition}\label{def:NormPossMCSEq}

Let $\overline M$ be a normal poss-MCS and $\overline{\mathcal{S}}$ a possibilistic belief state. $\overline{\mathcal{S}}$ is a possibilistic equilibrium of $\overline M$ if $\overline{\mathcal{S}} = \overline{MD} (\overline M^{\mathcal{S}})$.
\end{definition}

Paper \cite{yin12} presents also a fixpoint theory for definite poss-MCS, which provides a way for computing the equilibrium for both definite and normal poss-MCS. 

\begin{example} \label{possMCSexample}
In a different version of example \ref{MCSexample} all agents use possibilistic logic programs to encode their knowledge and bridge rules, forming a possibilistic MCS $\overline M$. The three agents are modelled as contexts $\overline{c_1}$, $\overline{c_2}$ and $\overline{c_3}$, respectively, with knowledge bases:
\begin{equation*}
\begin{split} 
\overline{P_1} =  & \left \{ \begin{array}{l} sensors \leftarrow, [1],\: corba \leftarrow, [1], \\ distributedComputing \leftarrow corba, not\: centralizedComputing, [0.8] \end{array} \right \}  \\
\overline{P_2} =  & \{  profA, [1] \} \\
\overline{P_3} = & \{ ambientComputing \leftarrow ubiquitousComputing, [0.9] \}
\end{split} 
\end{equation*}

\noindent and bridge rules:
\begin{equation*}
\begin{split} 
\overline{B_1} = & \left \{ \begin{array}{l} (c_1 : centralizedComputing) \leftarrow  (c_2 : middleware),  [0.7],\\ 
(c_1 : distributedComputing) \leftarrow  (c_3 :ambientComputing), [0.6]  \end{array} \right \} \\
\overline{B_2} = & \{ (c_2 : middleware) \leftarrow (c_1 : corba), [0.9] \} \\
\overline{B_3} =&  \{ (c_3: ubiquitousComputing) \leftarrow   (c_1 : sensors), (c_2 : profB), [0.8] \}
\end{split} 
\end{equation*}
 
\noindent Rules or facts with degree $[1]$ indicate that the agent is certain about them, while rules with degree less than $[1]$ indicate uncertainty about whether the rule holds.

$\overline M$ is a normal poss-MCS. In order to compute its possibilistic equilibrium, we first have to compute its reduct with respect to $\mathcal{S}$, where $\mathcal{S}$ is the grounded equilibrium of $M$ (the classical projection of $\overline M$):

\begin{equation*}
\begin{split} 
\mathcal{S} = & \left ( \begin{array}{l} \{sensors,corba,centralizedComputing\},\\ \{profA,middleware\}, \emptyset \end{array} \right )
\end{split} 
\end{equation*}
The reduct of $\overline M$ with respect to $\mathcal{S}$, $\overline M ^{\mathcal{S}}$, is derived from $\overline M$ by replacing $\overline{P_1}$ with $\overline{P_1}^{\mathcal{S}}$: $$\overline{ P_1}^{\mathcal{S}}=  \{ sensors, [1],\: corba, [1],  distributedComputing \leftarrow corba,  [0.8] \} $$
The next step is to compute the necessity of each atom in $\mathcal{S}$. Following Definition \ref{def:PossDistribution}, $\pi_{\overline M}(\mathcal{S})=1$, as $\mathcal{S}$ is the grounded equilibrium of $M$. For  
$$\mathcal{S'} = (\{sensors,corba\}, \{profA,middleware\}, \emptyset)$$
it holds that $\pi_{\overline M}(\mathcal{S}')=1-max\{0.8,0.7\}=0.2$, while for  
$$\mathcal{S''} = (\{sensors,corba\}, \{profA\}, \emptyset)$$   
it holds that $\pi_{\overline M}(\mathcal{S}'')=1-max\{0.8,0.7,0.9\}=0.1$. Using Definition \ref{def:AtomPossNec}, the necessities of the atoms in $\mathcal{S}$ are: $\Pi_{\overline M}(sensors)= 1$, $\Pi_{\overline M}(cobra)= 1$, $\Pi_{\overline M}(centralizedComputing)= 0.8$, $\Pi_{\overline M}(profA)= 1$ and $\Pi_{\overline M}(middleware)= 0.9$. The possibilistic equilibrium of $\overline M$ is therefore:
\begin{equation*}
\begin{split} 
\overline{\mathcal{S}} = & \left ( \begin{array}{l} \{(sensors, [1]), (corba, [1]) (,centralizedComputing, [0.8])\},\\ \{(profA, [1]), (middleware, [0.9])\}, \emptyset \end{array} \right )
\end{split} 
\end{equation*}
 \end{example}

\section{Main Example }\label{sec:example}

We now present a scenario to illustrate how our approach works. 
Consider an office building, where robots assist human workers. As typically, there are not enough office supplies, such as cutters, glue, etc., for everyone, they have to be shared among the workers. 
Furthermore, as it is considered inefficient and unproductive for a worker to contact other colleagues and get supplies by themselves, the workers can submit requests to the robots to get and/or deliver the needed supplies for them, while they keep on working at their desks. We refer to a request submitted to the robots as a task.

\begin{figure}[h!]
		\includegraphics[scale=0.47] {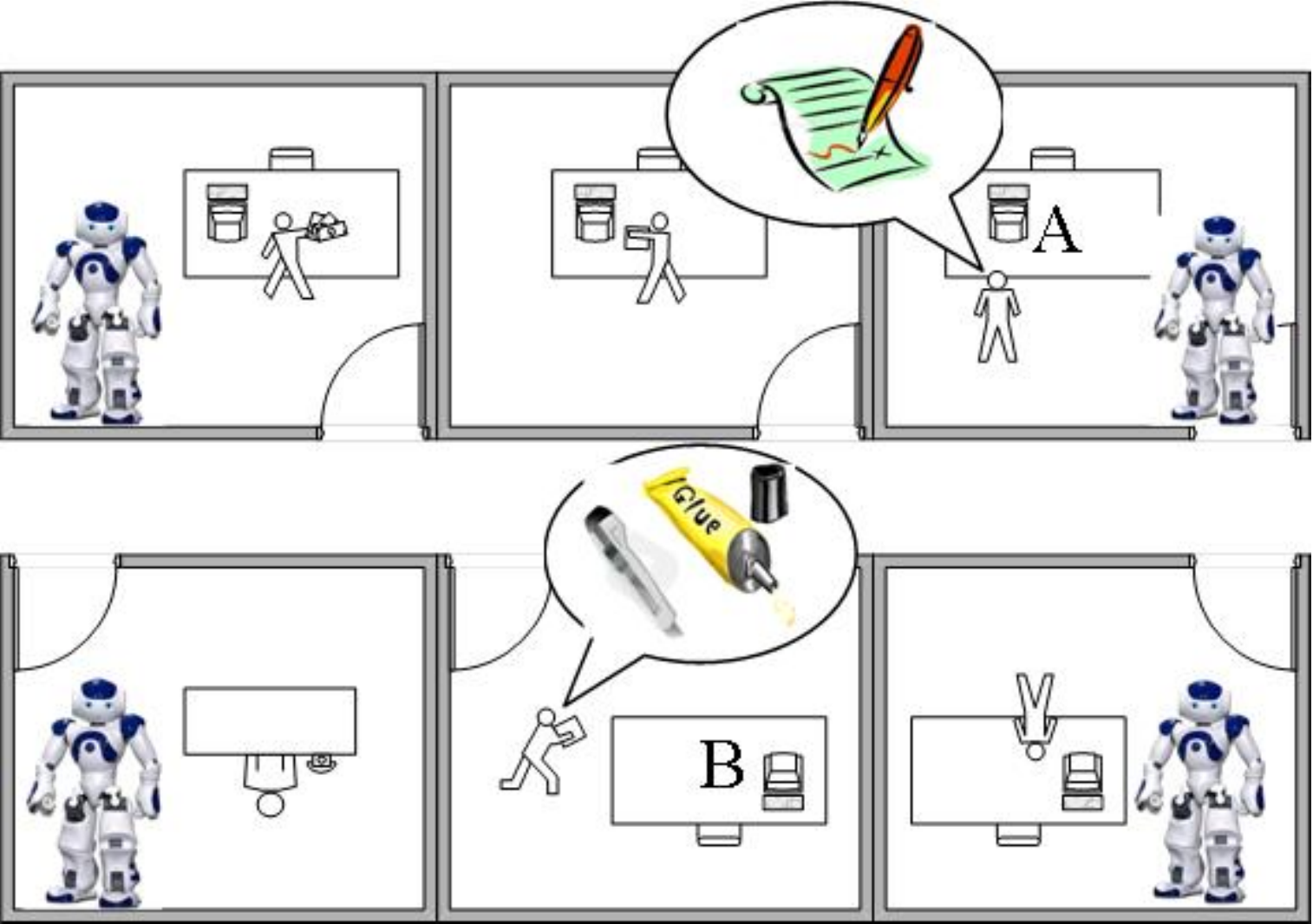}
		\caption{A depicted scenario of robots in office building.}
	\label{fig:nao-task}
\end{figure}

Workers and robots communicate via a simple web-based application, which transmits the workers' requests to the robots and keeps track of their status. The robots have limited computational resources: they only keep track of their recent past. Furthermore, not all robots know about the exact locations of supplies. Therefore, robots rely on each other for information about  the location of the supplies: the last robot having dealt with a supply is the one knowing where it is. We assume the availability of such an application, and a stable and reliable communication network. A depiction of the scenario is presented in Figure \ref{fig:nao-task}.

We consider a set of four robots $\mathcal{A}g=\{ag_1,a g_2, a g_3,$ $ag_4\}$ and four tasks: $\mathcal{T}= \{t_1, t_2,t_3, t_4 \}$, where
$t_1$ is to deliver a pen to desk $D_a$,  
$t_2$ is to deliver a piece of paper to desk $D_a$,
$t_3$ is to deliver a tube of glue to desk $D_b$, 
and $t_4$ is to deliver a cutter to desk $D_b$.
We assume that a robot can perform a task if it can carry the relevant material and knows its source and destination. Due to their functionalities, robots can carry the following material: $ag_1$ the pen or the glue, $ag_2$ the paper, $ag_3$ the glue or the cutter, and $ag_4$ the pen or the cutter. 
Each robot knows who has the information about the source and the destination of each material, but the actual coordinates are only revealed after an agreement on a coalition among the robots has been made. This involves interdependency among robots. 

To start, robots get the information concerning the locations of the supplies and the distances between the material and their destinations. Tables \ref{tab:naoknows} and \ref{tab:locations} present the knowledge of the robots about the tasks and the current distances among the robots, the material and the destinations, respectively. The table should be read as follows. Robot $ag_1$, regarding task $t_1$, knows nothing about the source of the pen, i.e., where it currently is, but does know the destination for the pen, i.e., where it must be delivered. Regarding task $t_2$, robot $ag_1$ knows where the paper is, but knows nothing about its destination. 
 
\begin{table}[h!]
\caption{\label{tab:naoknows} Robots' knowledge and capabilities}
\begin{tabular}{|c|c|c|c|c|c|c|c|c|}
\hline
Robot		
	& \multicolumn{4}{c|}{$ag_1$}
	& \multicolumn{4}{c|}{$ag_2$}
	\\ 
\hline
Task		
	& $t_1$			
	& $t_2$
	& $t_3$
	& $t_4$	
	& $t_1$
	& $t_2$	
	& $t_3$
	& $t_4$
	\\ 
\hline
Source 		
	&
	& x 
	&
	&
	& x 
	&
	&
	&
	\\ 
\hline
Destination 
	& x 
	&
	& x 
	&
	&
	&
	& 
	& x 
	\\ 
\hline 
\hline
Robot
	& \multicolumn{4}{c|}{$ag_3$}
	& \multicolumn{4}{c|}{$ag_4$}
	\\ 
\hline
Task
	& $t_1$	
	& $t_2$
	& $t_3$
	& $t_4$
	& $t_1$
	& $t_2$
	& $t_3$
	& $t_4$
	\\ 
\hline
Source
	&
	&
	&
	& x 
	&
	&
	& x 
	&
	\\ 
\hline
Destination
	&
	& x 
	&
	& 
	&
	& 
	&
	&
	\\ 
\hline
\end{tabular}

\caption{\label{tab:locations} Distances among locations}
\begin{tabular}{|c|c|c|c|c|}
\hline
\multicolumn{5}{|c|}{Distances among locations} \\
\hline
Robot	&	Pen		& Paper		&	Glue		& Cutter	\\
\hline
$ag_1$	&	10		&	15			&	9				& 12			\\
$ag_2$	&	14		&	8				&	11			&	13			\\
$ag_3$	&	12		&	14			&	10			&	7				\\
$ag_4$	&	9			&	12			&	15			& 11			\\
\hline
\hline
Destination		& Pen & Paper & Glue & Cutter \\
\hline		
$D_a$	&	11	&	16	&	9		&	8 \\
$D_b$	&	14	&	7		&	12	&	9	\\
\hline
\end{tabular}

\end{table}

Upon receiving information about the tasks, robots generate plans to carry out the tasks based on their knowledge and capabilities. For example, there are two different plans for delivering the pen to desk $D_a$ ($t_1$): $ag_1$ can deliver it after receiving information about its  location from robot $ag_2$; alternatively, $ag_4$ can deliver it after receiving information about its location from $ag_2$ and about its destination from $ag_1$. Given the plans, the robots then need to decide how to form coalitions to execute the tasks. We refer to a coalition as a group of robots executing a task. For example to accomplish all tasks $t_1, t_2, t_3, t_4$, the following two coalitions may be formed:
\begin{equation*}\label{eq:bridge}
\begin{split}
C_0:  = & \{(ag_1, t_3),(ag_2, t_2),(ag_3, t_4),(ag_4, t_1)\} \\
C_1:  = & \{(ag_1, t_1),(ag_2, t_2),(ag_3, t_3),(ag_4, t_4)\} \\
\end{split}
\end{equation*}
After forming coalitions, each robot has to generate its own plan to carry out the assigned tasks, e.g. plan the optimal route to get the material and carry it to its destination. 
%
%

Typically, route planning programs split into two main parts, the representation of the environment and a method of searching possible route paths between the current robot position and some new location, avoiding the obstacles which are known. Hence, mobile robot navigation planning requires to have sufficient reliable estimation of the current location of the robot, and also a precise map of the navigation space. Path planning takes into consideration a model or a map of the environment or contexts, to determine what are the geometric path points for the mobile robots to track from a start position to the goal to be reached. 

The most commonly used algorithms for these methods are the $A^*$ algorithm, a global search algorithm giving a complete and optimal global path in static environments, and its optimization the $D^*$ algorithm. Other examples in the literature include using distributed route planning methods for multiple mobile robots, using the Lagrangian decomposition technique, neural networks \cite{DBLP:journals/cor/SomhomME99} and genetic algorithms \cite{CaiPeng}. One of the lessons which has been learned in this research area is that the need for optimal planning is outweighed by the need for quickly finding an appropriate plan \cite{DBLP:journals/trob/KoenigL05}.

In this paper our focus is on finding and selecting among the possible coalitions with which a given set of goals will be reached, rather than on the individual plans of the agents to carry out their assigned tasks.

%

\section{Computing and evaluating coalitions in a perfect world}\label{sec:compute}

One question that arises in scenarios such as the one that we present in Section \ref{sec:example} is how to compute the alternative coalitions that may be formed to achieve a set of given goals. Here we present a solution based on the use of heterogeneous nonmonotonic MCS \cite{brewkaEiter07}, described in Section \ref{sec:background}. The main reasons for choosing the MCS model are: (a) it enables representing systems consisting of agents with different knowledge representation models; (b) it can represent different kinds of relationships among agents such as goal-based dependencies, constraints and conflicting goals; and (c) it provides both centralized and distributed reasoning algorithms, which can be used for computing goal-based coalitions. Roughly, our solution consists in representing agent dependencies and inter-agent constraints using \emph{bridge rules} and computing the potential coalitions using algorithms for MCS equilibria.

\subsection{Modeling dependencies}

We model each agent in a multiagent system as a context in a MCS. The knowledge base of the context describes the goals of the agent and the actions that it can perform. Goals and actions are represented as literals of the form $g_k$, $a_j$, respectively. Bridge rules represent the dependencies of the agent on other agents to achieve its goals. According to the definition given by \cite{Sichman01onsocial}, a dependence relation 
$$dp:basic\_dep(ag_i,ag_j,g_k,p_l,a_m)$$ denotes that agent $ag_i$ depends on agent $ag_j$ to achieve goal $g_k$, because $ag_j$ can perform action $a_m$ needed in the plan $p_l$, which achieves the goal; for a goal $g_k$ of agent $ag_i$, which is achieved through plan $p_l=(ag_1:a_1,ag_2:a_2,...,ag_n:a_n)$, where $ag_j:a_j$ represents action $a_j$ performed by agent $ag_j$, the following dependence relations hold:
$$dp_j:basic\_dep(ag_i,ag_j,g_k,p_l,a_j), j=\{1,...,n\} $$
We denote this set of dependencies as $DP(ag_i,g_k,p_l)$. One way to represent dependencies is by using rules of the form: $Head \leftarrow Body$, where the $Head$ denotes the goal of agent $ag_i$ that is to be achieved ($g_k$), and the $Body$ describes the actions of plan $p_l$ that will lead to the achievement of the goal. Based on this intuition, we define bridge rules representing dependence relations among agents as follows:
\begin{definition}
\label{def:BR}
For an agent $ag_i$ with goal $g_k$ achieved through plan $p_l=(ag_1:a_1,ag_2:a_2,...,ag_n:a_n)$, the set of dependencies $DP(ag_i,g_k,p_l)$ is represented by a bridge rule of the form:
\begin{equation}
(c_i:g_k) \leftarrow (c_1:a_1),(c_2:a_2),...,(c_n:a_n)
\end{equation}
where $c_j$, $j=1,...,i,...,n$ is the context representing agent $ag_j$. 
\end{definition}
Based on the above representation of agents as contexts, and goal-based dependencies among agents as bridge rules, we represent multiagent systems as MCS as follows.
\begin{definition}
\label{def:MCS}
A MCS $M(\mathcal{A})$ corresponding to a multiagent system $\mathcal{A}$ is a set of contexts $c_i=(L_i,kb_i,br_i)$, where \emph{L$_i$} = (\textbf{KB}$_i$, \textbf{BS}$_i$, \textbf{ACC}$_i$) is the logic of agent $ag_i\in \mathcal{A}$, $kb_i \in \textbf{KB}_i$ is a knowledge base that describes the actions that $ag_i$ can perform and its goals, and $br_i$ is a set of bridge rules, a subset of which represents the dependencies $DP(ag_i,g_k,p_l)$ of $ag_i$ on other agents in $\mathcal{A}$ for all goals $g_k$ of $ag_i$ and all plans $p_l$, with which these goals can be achieved.
\end{definition}

The main advantage of this model is that it enables agents using different logics to describe their actions and goals, to form plans cooperatively by exchanging information through their bridge rules. Assuming a signature $\Sigma$, the following are some example logics that are captured by definition \ref{def:MCS}:

\begin{itemize}
\item Default logic \cite{reiter80}: \textbf{KB} is the set of default theories based on $\Sigma$;  \textbf{BS} is the set of deductively closed sets of $\Sigma$-formulas; and \textbf{ACC}($kb$) is the set of $kb$'s extensions.
\item Normal logic programs under answer-set semantics \cite{gelfond91}: \textbf{KB} is the set of normal logic programs over $\Sigma$;  \textbf{BS} is the set of sets of atoms over $\Sigma$; and \textbf{ACC}($kb$) is the set of $kb$'s answer sets.
\item Propositional logic under the closed-world assumption: \textbf{KB} is the set of sets of propositional formulas over $\Sigma$;  \textbf{BS} is the set of deductively closed sets of propositional $\Sigma$-formulas; and \textbf{ACC}($kb$) is the set of $kb$'s consequences under the closed world assumption.
\end{itemize}

There are numerous other examples, both monotonic (e.g., description logics, modal logics,
temporal logics), and nonmonotonic (e.g., circumscription, autoepistemic logic, defeasible logic). 

This feature (the generality of the representation model) is particularly important in open environments, where agents are typically heterogenous with respect to their representation and reasoning capabilities (e.g. Ambient Intelligence systems).

\begin{example} \label{ex:mainExample}
In our main example, introduced in Section \ref{sec:example}, we assume that all four robots use propositional logic. We model the four robots, $ag_1$-$ag_4$, as contexts  $c_1$-$c_4$, respectively, with the following knowledge bases:
%
\begin{equation*}
\begin{split}
kb_1 = & \{a_{2s},a_{1d},a_{3d},a_{1c} \vee a_{3c}\}\\
kb_2 = & \{a_{1s},a_{4d},a_{2c}\}\\
kb_3 =  & \{a_{4s},a_{2d},a_{3c} \vee a_{4c}\}\\
kb_4 =  & \{a_{3s},a_{1c} \vee a_{4c}\} 
\end{split}
\end{equation*}

\noindent where $a_{ij}$ represents the actions that a robot can perform. $i$ stands for the object to be delivered: $1$ stands for the pen, $2$ for the paper, $3$ for the glue and $4$ for the cutter. $j$ stands for the kind of action that the agent can perform: $c$ stands for carrying the object, $s$ stands for providing information about the current location (source) of the object, while $d$ stands for providing information about the destination of the object.  For example, $ag_1$ can 

\begin{itemize}
\item provide information about the source of the paper ($a_{2s}$)
\item provide information about the destination of the pen ($a_{1d}$)
\item provide information about the destination of the glue ($a_{3d}$)
\item carry the pen or the glue ($a_{1c} \vee a_{3c}$)
\end{itemize}

We represent the four tasks that the robots are requested to perform, $t_i$, as goals, $g_i$. For example $g_1$ represents the task of delivering the pen to desk $D_a$ ($t_1$). We also assume that a robot $ag_j$ can fulfil goal $g_i$, i.e. deliver object $i$ to its destination, if it can perform action $a_{ic}$, i.e. carry object $i$. For example, $g_1$ can be fulfilled by robots $ag_1$ and $ag_4$, because these robots  can carry the pen ($a_{1c}$). 

Given the knowledge and capabilities of robots, as described in Table \ref{tab:naoknows}, the robots can fulfil goals $g_1-g_4$ as follows. For $g_1$, there are two alternative plans:
\begin{equation*}
\begin{split}
p_{11} = &(ag_2:a_{1s},ag_1:a_{1c})\\
p_{12} = &(ag_2:a_{1s},ag_1:a_{1d},ag_4:a_{1c})
\end{split}
\end{equation*}
According to $p_{11}$, robot $ag_2$ must provide information about the source of the pen ($ag_2:a_{1s}$) and $ag_1$ must carry the pen to its destination ($ag_1:a_{1c}$). According to $p_{12}$, robot $ag_2$ must provide information about the source of the pen ($ag_2:a_{1s}$), $ag_1$ must provide information about its destination ($ag_1:a_{1d}$), and $ag_4$ must carry the pen to its destination ($ag_4:a_{1c}$). 

For $g_2$ there is only one plan, $p_{21}$; for $g_3$ there are two alternative plans: $p_{31}$ and $p_{32}$; and for $g_4$ there are two plans: $p_{41}$ and $p_{42}$:
\begin{equation*}
\begin{split}
p_{21} = &(ag_1:a_{2s},ag_3:a_{2d},ag_2:a_{2c})\\
p_{31} = &(ag_4:a_{3s},ag_1:a_{3c})\\
p_{32} = &(ag_4:a_{3s},ag_1:a_{3d},ag_3:a_{3c})\\
p_{41} = &(ag_2:a_{4d},ag_3:a_{4c})\\
p_{42} = &(ag_3:a_{4s},ag_2:a_{4d},ag_4:a_{4c})\\
\end{split}
\end{equation*}
Each plan implies dependencies among robots. For example, from $p_{11}$ the following dependency is derived: $$dp_1:basic\_dep(ag_1,a g_2,g_1,p_{11},a_{1s})$$
namely $ag_1$ depends on $ag_2$ to achieve goal $g_1$, because $ag_2$ can provide information about the source of the pen ($a_{1s}$). Figure \ref{fig:modelAll} represents the dependencies derived from all plans, abstracting from plans, similarly to \cite{DBLP:conf/atal/SichmanC02}. The figure should be read as follows: The pair of arrows going from node $ag_1$ to the rectangle box labeled $a_{1s}$ and then to node $ag_2$ indicates that agent $ag_1$ depends on agent $ag_2$ to achieve goal $g_1$, because the latter can perform action $a_{1s}$. 

\begin{figure} [h!] 
		\includegraphics[scale=0.75] {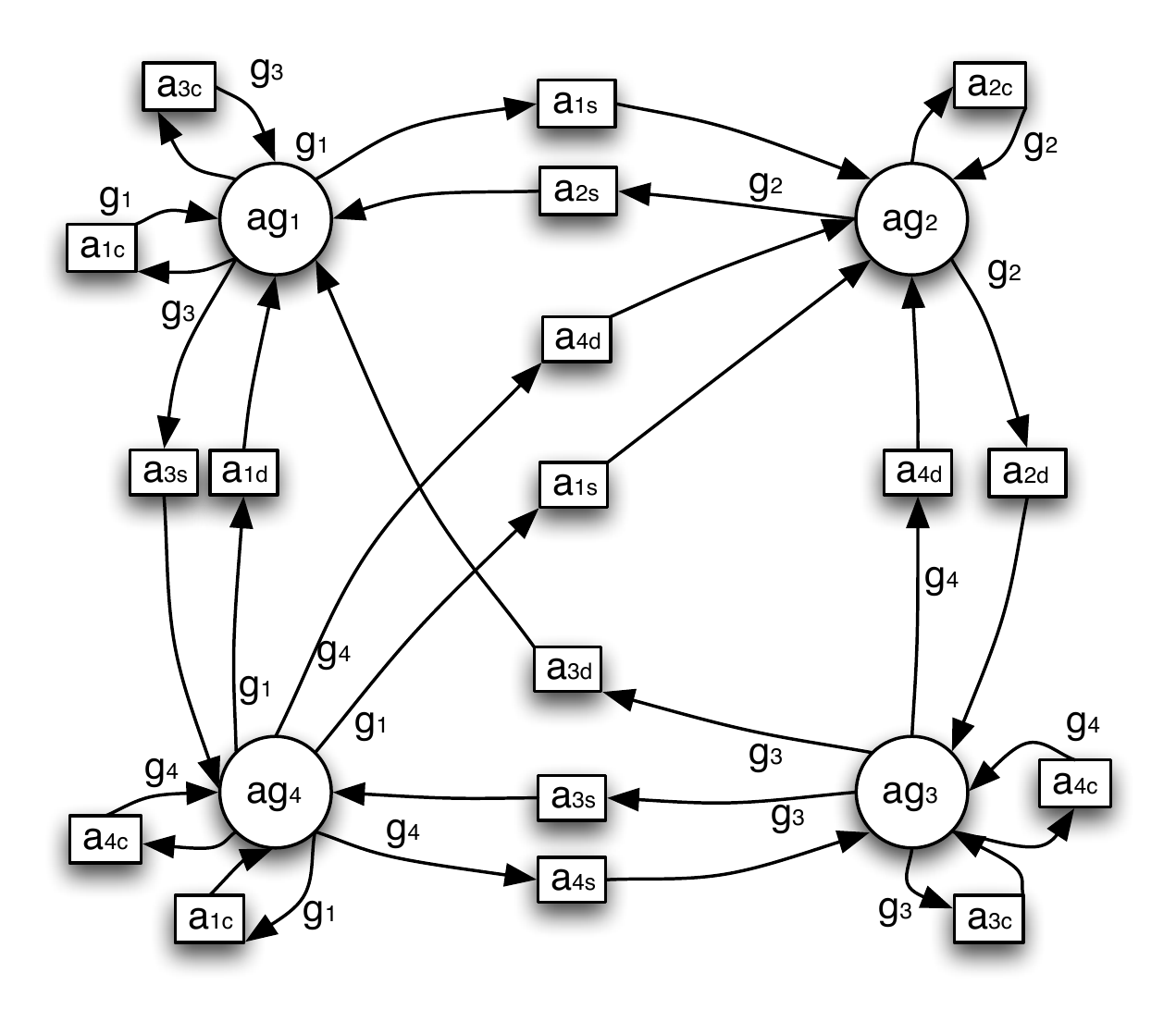}
		\vspace{-12pt}
		\caption{Dependencies among the four robot agents.}
	\label{fig:modelAll}
\end{figure}
Bridge rules $r_1$-$r_7$ represent the same dependencies. Each rule represents the dependencies derived by a different plan. For example $r_1$ corresponds to plan $p_{11}$ and represents dependency $dp_1$. 
\begin{equation*}
\begin{split}
r_1=(c_1:g_1) & \leftarrow  (c_1:a_{1c}),(c_2:a_{1s}) \\
r_2=(c_4:g_1) & \leftarrow  (c_4:a_{1c}),(c_2:a_{1s}),(c_1:a_{1d})\\ 
r_3=(c_2:g_2) & \leftarrow  (c_2:a_{2c}),(c_1:a_{2s}),(c_3:a_{2d})\\
r_4=(c_1:g_3) & \leftarrow  (c_1:a_{3c}),(c_4:a_{3s}) \\
r_5=(c_3:g_3) & \leftarrow  (c_3:a_{3c}),(c_4:a_{3s}),(c_1:a_{3d})\\ 
r_6=(c_3:g_4) & \leftarrow  (c_3:a_{4c}),(c_2:a_{4d}) \\
r_7=(c_4:g_4) & \leftarrow  (c_4:a_{4c}),(c_3:a_{4s}),(c_2:a_{4d})
\end{split}
\end{equation*} 
One system constraint is that two robots cannot carry the same object at the same time. In order to represent this constraint, we expand the body of each of the bridge rules with a predicate that describes the fact that another robot carries the material that the rule refers to:
\begin{equation*}
\begin{split}
r_1'=(c_1:g_1) & \leftarrow  (c_1:a_{1c}),(c_2:a_{1s}), not(c_1:carriesElse_1)\\
r_2'=(c_4:g_1) & \leftarrow  (c_4:a_{1c}),(c_2:a_{1s}),(c_1:a_{1d}), not(c_4:carriesElse_1)\\ 
r_3'=(c_2:g_2) & \leftarrow  (c_2:a_{2c}),(c_1:a_{2s}),(c_3:a_{2d}), not(c_2:carriesElse_2)\\
r_4'=(c_1:g_3) & \leftarrow  (c_1:a_{3c}),(c_4:a_{3s}), not(c_1:carriesElse_3) \\
r_5'=(c_3:g_3) & \leftarrow  (c_3:a_{3c}),(c_4:a_{3s}),(c_1:a_{3d}), not(c_3:carriesElse_3)\\ 
r_6'=(c_3:g_4) & \leftarrow  (c_3:a_{4c}),(c_2:a_{4d}), not(c_3:carriesElse_3) \\
r_7'=(c_4:g_4) & \leftarrow  (c_4:a_{4c}),(c_3:a_{4s}),(c_2:a_{4d}), not(c_4:carriesElse_4)
\end{split}
\end{equation*}
For example $c_1:carriesElse_1$, which appears in $r_1'$, represents that an agent different to $ag_1$ carries material $1$ (the pen).

For each of the four robots $ag_l$, and for each material $i$ that the robot can carry, we also add bridge rules of the form: $$(c_l:carriesElse_i) \leftarrow (c_k:a_{ic})$$ where $i,k,l=\{1...4\}$ and $k\neq l$, which describe the cases that an agent different to $ag_l$ carries material $i$. For example, the following rules describe the cases that a robot different to $ag_1$ carries the pen.
\begin{equation*}
\begin{split}
r_8'=(c_1:carriesElse_1)& \leftarrow (c_2:a_{1c}) \\
r_9'=(c_1:carriesElse_1)& \leftarrow  (c_3:a_{1c}) \\
r_{10}'=(c_1:carriesElse_1)& \leftarrow  (c_4:a_{1c}) 
\end{split}
\end{equation*}
\end{example}

We should note here that the same example can also be represented using dynamic MCS \cite{dao-tran11}, which use schematic bridge rules that are instantiated at run time. For example, using dynamic MCS, we would replace the static predicates $(c_1:carriesElse_1)$ in rule $r_1'$ and $(c_4:carriesElse_1)$ in $r_2'$ by a dynamic predicate $(c_x:a_{1c})$, which would be instantiated at run time by any context $c_x$, which contains $a_{1c}$ in its knowledge base; in other words by any robot that can carry material $1$. Rules such as $r_8'$, $r_9'$ and $r_{10}'$ would then be unnecessary and would be omitted. Such an approach is more appropriate for open and dynamic environments, where any agent may join or leave the system at any time and without prior notice (e.g. Ambient Intelligence environments).

In the running example, we assumed (for reasons of simplicity) that the four agents use propositional logic. However, with MCS it is possible to represent any agent using a logic that can be captured by Definition \ref{def:MCS}. Note also that we use a rather simplistic representation for plans, because our goal is not to represent and reason with plans; we are only interested in the dependencies derived from plans.

\subsection{Computing coalitions}

An equilibrium in MCS represents an acceptable belief state of the system. Each belief set in this state is derived from the knowledge base of the corresponding context and is compatible with the applicable bridge rules. For a MCS $M(\mathcal{A})$ that corresponds to a multiagent system $\mathcal{A}$, an equilibrium $\mathcal{S}=\{S_1,...,S_n\}$ represents a coalition in which agents of $\mathcal{A}$ can achieve their goals. Specifically, each belief set $S_i$ in the equilibrium contains the actions that agent $ag_i$ can perform and the goals that it will achieve in this coalition. If there is more than one ways with which the goals can be achieved, the MCS will have more than one equilibria, each one representing a different coalition. If a certain goal does not appear in any of the equilibria, this means that there is no coalition with which the goal can be achieved.

In order to compute the potential coalitions in a multiagent system $\mathcal{A}$, one then has to formulate the MCS $M(\mathcal{A})$ that corresponds to $\mathcal{A}$, and compute the equilibria $\mathcal{S}$ of $M(\mathcal{A})$. 

Equilibria in nonmonotonic MCS can be computed using any of the following algorithms / implementations depending on the requirements of the specific system or application, e.g. with respect to the representation and reasoning capabilities of the agents:

\begin{itemize}
\item The MCS-IE system \cite{bogl10} implements a centralized reasoning approach that is based on the translation of MCS into HEX-programs \cite{eiter05} (an extension of answer set programs with external atoms), and on their execution in the dlv-hex system\footnote{\url{http://www.kr.tuwien.ac.at/research/systems/dlvhex/}}.
\item The algorithms proposed in \cite{bikakisKAIS11} implement a distributed computation, which however assumes that all contexts are homogeneous with respect to the logic that they use (defeasible logic).
\item The three algorithms proposed in \cite{dao-tran15} enable distributed computation of equilibria. DMCS assumes that each agent has minimal knowledge about the world, namely the agents that it is connected through the bridge rules, but doesn't have any further metadata, e.g. topological information, of the system. Its computational complexity is exponential to the number of literals used in the bridge rules. DMCS-OPT uses graph theory techniques to detect any cycle dependencies in the system and avoid them during the evaluation of the equilibria, improving the scalability of the evaluation. DMCS-STREAMING computes the equilibria gradually ($k$ equilibria at a time), reducing the memory requirements for the agents. The three algorithms have been implemented in a system prototype\footnote{\url{http://www.kr.tuwien.ac.at/research/systems/dmcs}}.
\end{itemize} 

Paper \cite{dao-tran15} presents a performance evaluation of the algorithms it proposes, as well as a comparison with previous approaches. They conclude that the algorithms in \cite{bikakisKAIS11} are in general much faster as they are based on a low-complexity logic, while among the others  there is no clear winner, as their performance depends on the memory capacity of the agents, and the topology of the system. 

Choosing the best approach for the computation of coalitions depends on several parameters, such as the targeted environment, the available means of communication, the computational, communication and knowledge representation capabilities of the agents, and the specific needs and requirements of the use cases that we want to support. For small-scale systems, such as the one in our running example, a centralized approach, e.g. the one proposed in \cite{bogl10}, is probably more appropriate as it achieves better performance when the total number of agents is small. For larger-scale systems, a distributed scalable approach is probably more appropriate. In that case, if the main requirement is to compute all possible coalitions as fast as possible, DMCS and DMCS-OPT should be preferred; DMCS for systems with simple topology, i.e. fewer dependencies among the agents, and DMCS-OPT for more complex systems with possible cycle dependencies among the agents. Finally, in cases that the memory capacity of the available agents is limited, or that we are only interested in computing fast some (and not all) of the possible coalitions, DMCS-STREAMING seems to be the most appropriate approach.

\begin{example}
Returning to our main example, the MCS that corresponds to the system of the four robots, $M(\mathcal{A})$, has two equilibria: $\mathcal{S}^0$ and $\mathcal{S}^1$:

\begin{equation*}
\begin{split}
\mathcal{S}^0= & \left ( \begin{array}{l} \{a_{2s},a_{1d},a_{3d},a_{3c},g_3\}, \\ 
\{a_{1s},a_{4d},a_{2c},g_2\}, \\ 
\{a_{4s},a_{2d},a_{4c},g_4\}, \\
\{a_{3s},a_{1c},g_1\}
\end{array} \right ) \\
\mathcal{S}^1 = & \left ( \begin{array}{l} \{a_{2s},a_{1d},a_{3d},a_{1c},g_1\}, \\ 
\{a_{1s},a_{4d},a_{2c},g_2\}, \\ 
\{a_{4s},a_{2d},a_{3c},g_3\}, \\
\{a_{3s},a_{4c},g_4\}
\end{array} \right )
\end{split}
\end{equation*} 

$\mathcal{S}^0$ represents coalition $C_0$, according to which $ag_1$ delivers the glue to desk $D_b$ ($g_3$), $ag_2$ delivers the paper to desk $D_a$ ($g_2$), $ag_3$ delivers the cutter to desk $D_b$ ($g_4$) and $ag_4$ delivers the pen to desk $D_a$ ($g_1$). $\mathcal{S}^1$ represents coalition $C_1$, according to which $ag_1$ delivers the pen to desk $D_a$ ($g_1$), $ag_2$ delivers the paper to desk $D_a$ ($g_2$), $ag_3$ delivers the glue to desk $D_b$ ($g_3$) and $ag_4$ delivers the cutter to desk $D_b$ ($g_4$). Using the previous abstraction of plans, the two coalitions are graphically represented in Figure \ref{fig:S0S1}. 

\begin{figure} [h!] 
		\includegraphics[scale=0.47] {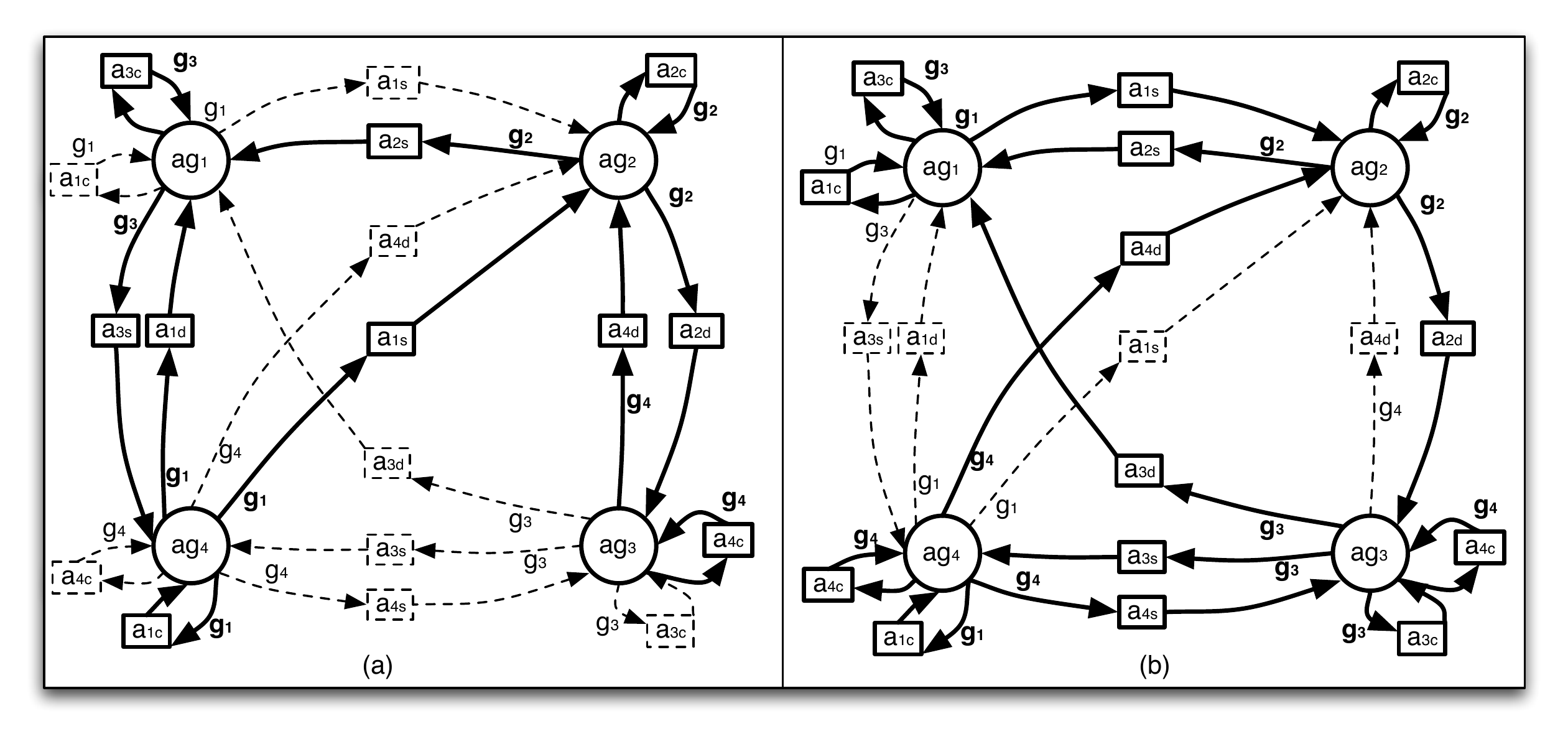}
		\caption{Coalitions $C_0$ (a), and $C_1$ (b) in bold; remaining dependencies in dotted lines. }
	\label{fig:S0S1}	
\end{figure}
\end{example}

In order to achieve their goals, the robots then have to carry out the actions in the plans that are associated to these goals. For example, for coalition $C_0$ the associated plans are: $p_{12}$ (for goal $g_1$), $p_{21}$ (for $g_2$), $p_{31}$ (for $g_3$) and $p_{41}$ (for $g_4$), while the plans associated to $C_1$ are $p_{11}$, $p_{21}$, $p_{32}$ and $p_{42}$.

\subsection{Selecting the best coalition}\label{sec:selectCoal}

Selecting the best among the possible coalitions in which agents can achieve their goals requires evaluating and comparing them. Efficiency and stability metrics are commonly used to evaluate coalitions (e.g., \cite{ecoeff,sha53,sch69,kl07}). The former giving an assurance on the economical gain reached by being in the coalition, the later giving a certainty that the coalition is viable on the long term. 

Generally speaking, efficiency in a coalition is a relation between what agents can achieve as part of the organization compared to what they can do alone or in different coalitions. Furthermore, a coalition is economically efficient iff i) no one can be made better off without making someone else worse off, ii) no additional output can be obtained without increasing the amount of inputs, iii) production proceeds at the lowest possible per-unit cost \cite{ecoeff}. 

\begin{example} 
In our running example, we can associate efficiency to the distances that the four robots must cover to perform the required actions. From Table \ref{tab:locations}  we can compute the distance for each robot to do each task, and, by adding them up, the cost of executing tasks in a given coalition:

\begin{equation*}
cost(C)=  \sum \limits_{i=1}^4 dist(ag_i,j)+(dist(j,dest(j))
\end{equation*} 
where $dist(ag_i,j)$ denotes the distance between the robot ($ag_i$) and the material ($j$) it has to carry in coalition $C$, and $dist(j,dest(j))$ denotes the distance between the material ($j$) and its destination ($dest(j)$). Based on this formula, the costs of $C_0$ and $C_1$ are:

\begin{equation*}
\begin{split}
cost(C_0)= &  (9+12) + (8+16) + (7+9) + (9+11) = 81 \\
cost(C_1)= &  (10+11) + (8+16) + (10+12) + (11+9) =87
\end{split}
\end{equation*} 

If we compare $C_0$ and $C_1$, $C_0$ is more economically efficient as at least one agent is better off without making anyone worse off, all else being equal (the distance that $ag_3$ has to cover is $16$ in $C_0$ and $22$ in $C_1$, while all other robots have to cover the same distance in both coalitions).
 $C_0$ is also more cost efficient than $C_1$, as its overall cost is smaller.
 
Other approaches could take into consideration the time it takes for each robot to accomplish its task and select the best coalition, which would in this case, be the one which executes the fastest all its tasks. To do this we would simply include a table of task duration for each robot along with the distances and compute the times similarly as for previously done for the distances. Moreover, other variables could be included such as the resources being actually used by each robot to accomplish its task, and the potential obstacle to avoid in order to reach their goals.

\end{example}

Stability of coalitions is related to the potential gain in staying in the coalition or quitting the coalition for more profit (i.e., free riding). Hence, several elements come to play for the evaluation of a coalition's stability. First, the coalition outcome should be greater than the individual ones cumulated. This is usually computed via a characteristic function such as the one proposed in \cite{mg92}. Second, the distribution of benefits should be fair. Several functions, named \emph{sharing rules} have been proposed for this purpose, such as the Shapley value \cite{sha53}, nucleolus \cite{sch69} and Satisfactory Nucleolus \cite{kl07}. The leading idea is to take the individual contribution
and the free rider's value into account when sharing the benefits.

Depending on the application domain, other functional and non-functional requirements, e.g., security, user-friendliness or conviviality, may also play an important role in the choice of a coalition. In \cite{caire:aamas11}, we compared coalitions in terms of \emph{conviviality}. Defined by Illich as ``individual freedom realized in personal interdependence" \cite{Illich71}, conviviality was introduced as a social science concept for multiagent systems to highlight soft qualitative requirements like user friendliness of systems. In \cite{caire:aamas11}, we proposed measures for conviviality to evaluate coalitions in dependence networks based on the following rough idea: more opportunities to work with others increases the system's conviviality. Specifically, we measured conviviality by the number of reciprocity-based coalitions that can be formed within an overall coalition. In our measures we used the notion of \emph{cycles} in the dependence graph, which denote the smallest graph topology expressing interdependence, and therefore conviviality. Given the dependence network ($DN$) that corresponds to a given coalition, the conviviality of the coalition $conv(DN)$ can be computed as follows:
\begin{eqnarray}\label{eq:globconv}
  \mathrm{conv}(DN) & =& \frac{\displaystyle\sum \mathrm{coal}(a,b)}{\Omega}, \\ [0.4em]
  \Omega 	& = & |\mathcal{A}|(|\mathcal{A}|-1) \times \Theta \label{eq:omega}, \\ [0.4em]
  \Theta 	& = & \sum_{l=2}^{l=|\mathcal{A}|} Perm(|\mathcal{A}|-2, l-2) \times |G|^{l} \label{eq:theta}, 
\end{eqnarray}
\noindent where $|\mathcal{A}|$ is the number of agents in the system, $|G|$ is the number of goals, $Perm$ is the usual permutation defined in combinatorics, \textbf{$\mathrm{coal}(a,b)$} for any distinct pair of agents $a,b \in \mathcal{A}$ is the number of cycles that contain the ordered pair ($a,b$) in $DN$, $l$ is the cycle length, and $\Omega$ denotes the maximal number of pairs of agents in cycles.

\begin{figure} [h!]
		\includegraphics[scale=0.50] {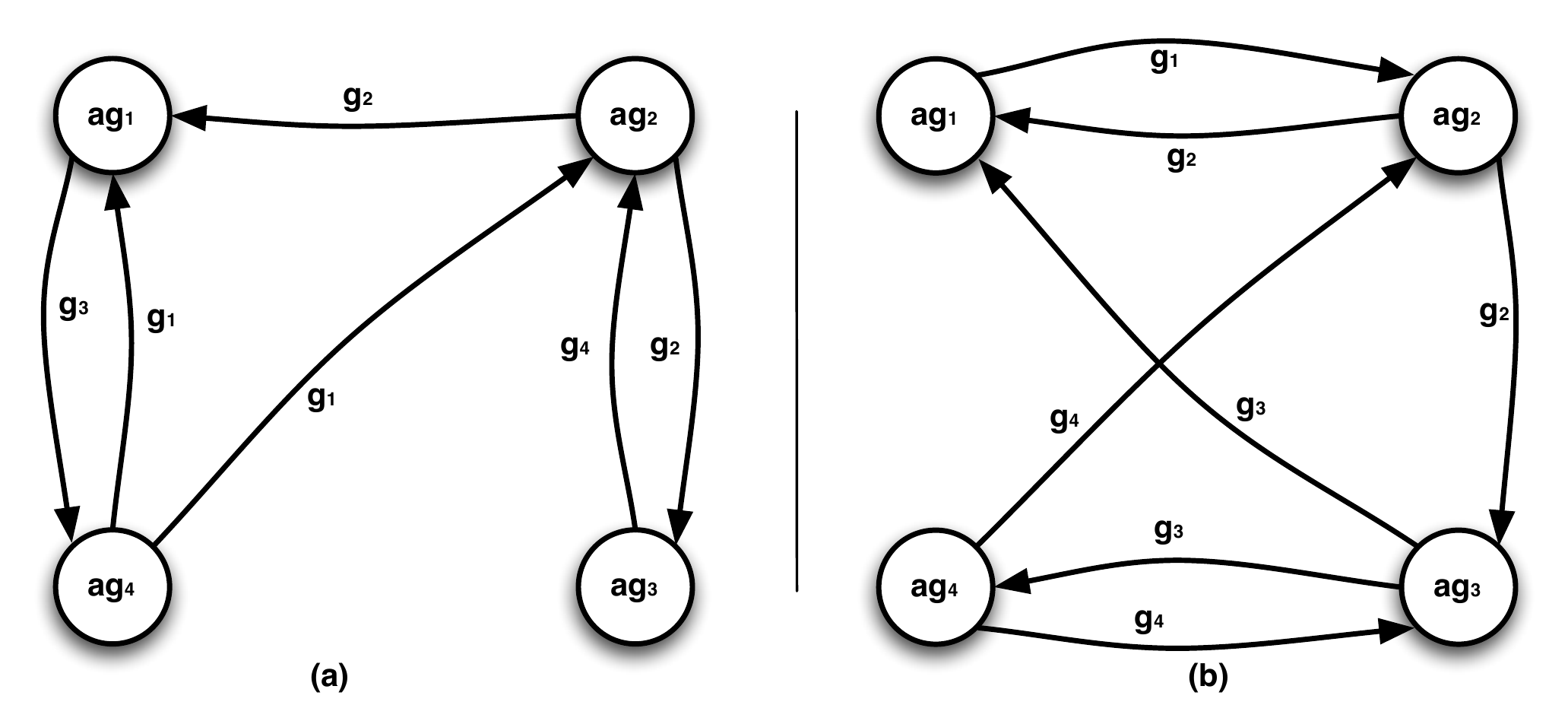}
		\caption{Goal dependencies in coalitions $C_0$ (a), and $C_1$ (b). }
	\label{fig:C0C1}
\end{figure}

\begin{example}
Back to our running example, abstracting from plans and actions, Figures \ref{fig:C0C1}.a and \ref{fig:C0C1}.b represent the dependence networks for coalitions $C_0$ and $C_1$ respectively. For both dependence networks, by applying formulae \ref{eq:omega} and \ref{eq:theta}, we get $\Theta = 656$ and $\Omega = 7872$. By applying formula \ref{eq:globconv}, we can then compute the conviviality of the two coalitions: $$conv(C_0)=0.00089, \: conv(C_1)=0.00127$$ 
Based on the above, $C_1$ is therefore preferred to $C_0$ in terms of conviviality.

\end{example}

In cases of agents with conflicting goals, coalitions differ in the set of goals that they can fulfil and the selection of a coalition depends on the priorities among the conflicting goals. It is among our future plans to integrate in the proposed model a preference relation on the set of goals to represent such priorities and develop algorithms for preference-based coalition formation.

The problem of selecting the best coalition however, has also been seen as tied to finding the optimal coalition structure, and therefore as a search problem which consists in performing a search on the coalition structure graph \cite{sandholm99}. A search through a subset of the coalition structure space is required: first, the bottom two levels of the coalition structure graph, then a time bounded breadth first search from the top of the coalition structure graph. The best coalition structure that is found is selected. This was optimised by \cite{DBLP:conf/atal/DangJ04}, who analysed fewer coalitions to establish small bounds from the optimal. In \cite{DBLP:journals/ai/ShehoryK98}, coalition sizes are limited and a greedy heuristic to yield a coalition structure is used; i.e. based on a limited number of agents.

\section{Computing coalitions under uncertainty} \label{sec:computeCoalitionUncertainty}

The models for representing and computing coalitions that we presented in Section \ref{sec:compute} are based on the \emph{perfect world assumption}, according to which it is always certain that agents will carry out the actions that are specified in the associated plans. However, in real-world settings, such assumption is not always valid. For instance, in our running example, there is uncertainty associated to the robots carrying the materials to their destinations. A robot may fall while on its way to pick up the material or towards its final destination; it may also run out of battery and fail to carry out its action. 

In this section we extend the representation and computational models that we presented in Section \ref{sec:compute} to take into account the uncertainty in the agents' actions. Similarly as in Section \ref{sec:compute}, we first present a rule-based representation model for dependencies, based on Possibilistic Multi-Context Systems \cite{yin12}. We then present the algorithms for computing the alternative coalitions and for selecting the best coalition under uncertainty using also multiple-criteria decision making methods.

\subsection{Modeling uncertainty}

We extend the definition of \cite{Sichman01onsocial} for dependence relations to take into account the uncertainty of actions:

\begin{definition} \label{dependUncertain}
The dependency of an agent $ag_i$ on agent $ag_j$ to achieve goal $g_k$, because $ag_j$ may perform action $a_m$ needed in plan $p_l$, which achieves the goal, is denoted as:
$$dp:poss\_dep(ag_i,ag_j,g_k,p_l,a_m,\alpha_{a_m})$$
where $\alpha_{a_m}$ represents the possibility that $ag_j$ will carry out action $a_m$ with success.

\end{definition}

We denote the set of dependencies of agent $ag_i$ to achieve goal $g_k$ through plan $p_l=(ag_1:a_1,ag_2:a_2,\ldots,ag_n:a_n)$ as $DP_{poss}(ag_i,g_k,p_l,\alpha_{p_l})$, where $\alpha_{p_l}$ represents the aggregated possibility that agents $ag_i,\ldots,ag_n$ will successfully perform  actions $a_1,\ldots,a_n$, respectively. Assuming that actions $a_1,\ldots,a_n$ are mutually independent: 
$$\alpha_{p_l}=\prod\limits_{i=1}^n \alpha_{a_i}$$

\noindent In the general case, where dependencies among the actions in $p_l$ may exist:
$$\alpha_{p_l}=(\prod\limits_{i=1}^{n-1} \alpha_{[a_i|a_{i+1}\land \ldots \land a_n]}) \times \alpha_{a_n}$$

\noindent where $\alpha_{[a_i|a_{i+1}\land \ldots \land a_{n}]}$ denotes the possibility that $a_i$ will be carried out successfully given that actions $a_{i+1}, \ldots ,a_{n}$ have been successful.

In order to model the uncertainty in the agents' actions, we model a multiagent system as a possibilistic MCS (or poss-MCS) with possibilistic bridge rules that represent the dependencies of the agent on the other agents to achieve its goals. 

\begin{definition}
\label{def:possBR}
For an agent $ag_i$ with goal $g_k$ achieved through plan $p_l=(ag_1:a_1,ag_2:a_2,...,ag_n:a_n)$, the set of dependencies $DP_{poss}(ag_i,g_k,p_l,\alpha_{p_l})$ is represented as a possibilistic bridge rule of the form:
$$c_i:g_k) \leftarrow (c_1:a_1),\ldots (c_n:a_n), [\alpha_{p_l}]$$
where $c_j$, $j=1,...,i,...,n$ is the context representing agent $ag_j$.
\end{definition}

Based on the above representations, we can now define the following representation for multiagents systems with uncertainty.
\begin{definition}
\label{def:possMC}
A poss-MCS $\overline M(\mathcal{A})$ corresponding to a multiagent system $\mathcal{A}$ with uncertainty is a set of possibilistic contexts $\overline c_i = (\Sigma _i,\overline P _i, \overline B _i)$, where $\Sigma _i$ is a set of atoms representing the goals of the agent and the actions it can perform; $\overline P _i$ represents the local theory of the agent as a possibilistic logic program; and $\overline B _i$ is a set of possibilistic bridge rules representing the dependencies of the agent on other agents to achieve its goals.
\end{definition}

As it becomes obvious, the above definition is more restrictive compared to definition \ref{def:MCS}, as it only allows agents that use possibilistic logic programs. However, by explicitly representing the degree of certainty in the rules, it offers a more elaborate manipulation of uncertainty, which is desirable when such values are available (e.g. in sensor-enriched environments). 

\begin{example} \label{ex:extendedExample}

In our running example, the uncertainty is associated with the robots carrying the materials to their destinations. In the robotics-related literature, different approaches may be found to compute uncertainty in (robotic) motion planning. Typically, uncertainty is considered in two ways: in sensing, where the current state of the robot and workspace may not be known with certainty, or in predictability, where the future state of the robot and workspace cannot be deterministically predicted even when the current state and future actions are known \cite{laValle2006}. Factors that may affect predictability are uncertainty in the workspace, e.g. moving obstacles \cite{vasquez2004}, uncertainty in the configuration of the robot \cite{hsiao2007}; or uncertainty in the robot's motion \cite{alterovitz2007}.

In our running example, the assumption is that uncertainty is associated with the motion of the robots and the grasping of a material. Furthermore, it is a function of the distance that the robot has to cover with and without the material. For our purpose, the following very simple function will suffice to compute the possibility that robot $ag_i$ carries out action $a_{jc}$; namely, carries material $c$ to its destination:
\begin{equation}\label{eq:robotPoss}
poss(ag_i,a_{jc})= 1-(0.001\times dist(ag_i,j)+0.002\times(dist(j,dest(j))
\end{equation}
where $dist(ag_i,j)$ denotes the distance between the robot ($ag_i$) and the material ($j$), and $dist(j,dest(j))$ denotes the distance between the material ($j$) and its destination ($dest(j)$). The intuition is that the uncertainty of a robot carrying a material to its destination is proportional to the distance that the robot has to cover to get to the material, and the distance that it has to cover after it grabs the material to deliver it to its destination. As many more things may go wrong while carrying the material, the second element has a double weight.

Using this function and the distances between the different locations as they appear in Table \ref{tab:locations}, the dependencies among the robots are represented by the following possibilistic bridge rules:
\begin{equation*}
\begin{split}
\overline{pr_1}=(c_1:g_1) & \leftarrow  (c_1:a_{1c}),(c_2:a_{1s}), not(c_1:carriesElse_1), [0.968] \\
\overline{pr_2}=(c_4:g_1) & \leftarrow  (c_4:a_{1c}),(c_2:a_{1s}),(c_1:a_{1d}),not(c_4:carriesElse_1), [0.969]\\ 
\overline{pr_3}=(c_2:g_2) & \leftarrow  (c_2:a_{2c}),(c_1:a_{2s}),(c_3:a_{2d}), not(c_2:carriesElse_2), [0.960]\\
\overline{pr_4}=(c_1:g_3) & \leftarrow  (c_1:a_{3c}),(c_4:a_{3s}), not(c_1:carriesElse_3), [0.967] \\
\overline{pr_5}=(c_3:g_3) & \leftarrow  (c_3:a_{3c}),(c_4:a_{3s}),(c_1:a_{3d}),not(c_3:carriesElse_3), [0.966]\\ 
\overline{pr_6}=(c_3:g_4) & \leftarrow  (c_3:a_{4c}),(c_2:a_{4d}), [0.975], not(c_3:carriesElse_3), \\
\overline{pr_7}=(c_4:g_4) & \leftarrow  (c_4:a_{4c}),(c_3:a_{4s}),(c_2:a_{4d}), not(c_4:carriesElse_4),[0.971]
\end{split}
\end{equation*} 
For example, the necessity degree of rule $\overline{pr_1}$ is associated with the uncertainty of action $(ag_1:a_{1c})$: robot $ag_1$ carrying the pen to desk $D_a$. According to function \ref{eq:robotPoss}, the possibility that this action will be carried out with success is: $$poss(ag_1,a_{1c})= 1-(0.001\times 10 +0.002\times 11)=0.968$$
where 10 is the distance between $ag_1$ and the pen, and 11 is the distance between the pen and desk $D_a$. As in the perfect world case, the negative atoms in the bridge rules (e.g. $not(c_1:carriesElse_1)$) represent that the material that a rule refers to (e.g. the pen) is not carried by another robot, and are derived from bridge rules such as the following:  
\begin{equation*}
\begin{split}
\overline{pr_8}=(c_1:carriesElse_1)& \leftarrow (c_2:a_{1c}), [1] \\
\overline{pr_9}=(c_1:carriesElse_1)& \leftarrow  (c_3:a_{1c}), [1]\\
\overline{pr_{10}}=(c_1:carriesElse_1)& \leftarrow  (c_4:a_{1c}), [1] 
\end{split}
\end{equation*}

\end{example}

\subsection{Computing coalitions under uncertainty}

In Section \ref{sec:background}, we showed that the possibilistic  equilibrium of a poss-MCS $\overline M$ is a collection of possibilistic atom sets $\overline S_i$, where each $\overline S_i$ is a collection of possibilistic atoms $\overline p_i$ from the system contexts. In the context of multiagent systems, an equilibrium represents a coalition in which agents can achieve their goals and $\overline p_i$ represents an action or a goal of agent $ag_i$ and its necessity degree. In the case of goals the necessity degree represents the level of certainty with which the coalition will achieve the goal.

To compute the potential coalitions under uncertainty in a multiagent system $\mathcal{A}$, one actually has to compute the possibilistic equilibria of $\overline M(\mathcal{A})$. This is possible using a fixpoint theory proposed in \cite{yin12}, which is based on the use of a consequence operator that gradually computes the consequences of a possibilistic MCS using its applicable rules.

\begin{example}
In our example, the poss-MCS that corresponds to the system of the four robots, $\overline M(\mathcal{A})$, has two possibilistic equilibria: $\overline{\mathcal{S}}^0$ and $\overline{\mathcal{S}}^1$
\end{example}
\begin{equation*}
\begin{split}
\overline{\mathcal{S}}^0= & \left ( \begin{array}{l} \{(a_{2s},[1]),(a_{1d},[1]),(a_{3d},[1]),(a_{3c},[0.967]),(g_3,[0.967])\}, \\ 
\{(a_{1s},[1]),(a_{4d},[1]),(a_{2c},[0.960]),(g_2,[0.960])\}, \\ 
\{(a_{4s},[1]),(a_{2d},[1]),(a_{4c},[0.975]),(g_4,[0.975])\}, \\
\{(a_{3s},[1]),(a_{1c},[0.969]),(g_1,[0.969])\}
\end{array} \right ) \\ \\
\overline{\mathcal{S}}^1= & \left ( \begin{array}{l} \{(a_{2s},[1]),(a_{1d},[1]),(a_{3d},[1]),(a_{1c},[0.968]),(g_1,[0.968])\},\\ 
\{(a_{1s},[1]),(a_{4d},[1]),(a_{2c},[0.960]),(g_2,[0.960])\},\\
\{(a_{4s},[1]),(a_{2d},[1]),(a_{3c},[0.966]),(g_3,[0.966])\},\\
\{(a_{3s},[1]),(a_{4c},[0.971],(g_4,[0.971])\}
\end{array} \right )
\end{split}
\end{equation*} 
$\overline{\mathcal{S}}^0$ represents coalition $C_0$ and $\overline{\mathcal{S}}^1$ represents coalition $C_1$; the two coalitions are graphically represented in Figure \ref{fig:S0S1}. $C_0$ will achieve goal $g_1$ with certainty degree 0.969, $g_2$ with 0.96, $g_3$ with 0.967 and $g_4$ with 0.975. On the other hand, $C_1$ will achieve goal $g_1$ with certainty degree 0.968, $g_2$ with 0.96, $g_3$ with 0.966 and $g_4$ with 0.971.

\subsection{Selecting the best coalition under uncertainty}

In Section \ref{sec:selectCoal} we presented different approaches for selecting the best coalition in the \emph{perfect world}, where there is certainty about the ability of the agents to carry out their tasks. If we take into account the uncertainty in the agents' actions, each coalition can achieve the agents' goals only with a certain degree of certainty. This is another criterion that we have to consider when evaluating the different coalitions.

The problem of selecting a coalition taking into account the degree of certainty with which the coalitions achieve the different goals is essentially a \emph{multi-attribute decision making} (MCDM) problem. We use the following definition for MCDM problems \cite{zimmermann91}:

\begin{definition}
\label{mcdm}
Let $D=\{D_i\}, i=1,\ldots,m$ be a set of decision alternatives and $R=\{R_j\}, j=1,\ldots,n$ a set of decision criteria according to which the desirability of an action is judged. Determine the optimal decision alternative $D^*$ with the highest degree of desirability with respect to all relevant decision criteria $R_j$.
\end{definition} 

In our case, the decision alternatives, $D$, are the alternative coalitions that the agents may form, and the criteria, $R$, are the certainty degrees with which the four goals will be reached. In the literature one can find many different alternatives for solving this problem, most of which are based on functions that aggregate the scores for the different criteria. Some characteristic examples are: 

\begin{itemize}
\item the \emph{Weighted Sum Method} \cite{fishburn67}, according to which each criterion $R_j$ is given a weight $w_j$, so that the sum of all weights is 1 ($\sum \limits_{j=1}^{n} w_j =1$), and the overall score of each alternative $D_i$ is the weighted sum of $q_{ij}$, i.e. the scores of $D_i$ for each criterion $R_j$:
$$WS(D_i) = \sum \limits_{j=1}^n w_j q_{ij}$$
The optimal decision alternative is the one with the highest overall score ($WS$). This method can only be used in single-dimensional cases, in which all the score units are the same. 

\item the \emph{Weighted Product Method} \cite{miller69}, which aggregates the individual scores using their product instead of their sum. Specifically, each decision alternative is compared with the others by multiplying a number of ratios, one for each criterion. Each ratio is raised to the power equivalent to the relative weight of the corresponding criterion. In general, in order to compare alternatives $D_1$ and $D_2$, the following product has to be calculated:
$$Ratio(D_1/D_2) = \prod \limits_{j=1}^{n}(q_{1j}/q_{2j})^{w_j}$$
where $q_{1j}$ and $q_{2j}$ are the scores of $D_1$ and $D_2$ respectively for criterion $R_j$, and $w_j$ is the weight of criterion $R_j$. If the ratio is greater than one, then $D_1$ is more desirable than $D_2$. The best  alternative is the one that is better, with respect to this ratio, than all other alternatives.  This method is sometimes called \emph{dimensionless analysis} because its structure eliminates any units of measure. It can, therefore, be used in both single- and multi-dimensional decision-making problems.

\item \emph{TOPSIS}  (\emph{Technique for Order Preference by Similarity to Ideal Solution} \cite{hwang81}), which is based on the concept that the chosen alternative should have the shortest geometric distance from the positive ideal solution and the longest geometric distance from the negative ideal solution. TOPSIS assumes that each alternative has a tendency of monotonically increasing or decreasing utility. Therefore, it is easy to locate the ideal and negative-ideal solutions. 
\end{itemize}

A review and comparative study of the most prominent methods for multi-criteria decision making is available in \cite{triantaphyllou2000}.

Computing coalitions in multiagent systems, as this is described in this paper, is a single-dimension problem, as the different criteria are the certainty degrees with which the goals are achieved, and are therefore of the same form and they take values in the same range ($[0,1]$). For such cases, the Weighted Product Method is considered as the preferred method because of its simplicity. 

\begin{example}
In our running example, there are two alternative coalitions with which the agents can achieve their goals: $C_0$ and $C_1$. If we apply the Weighted Sum Method, and considering that the four criteria represent the certainty degrees with which the four goals will be reached ($\alpha_{g_i}$), the weighted sum of a coalition is:
\end{example}
\begin{equation*}
WS(C)= \sum \limits_{i=1}^4 w_{g_i} \alpha_{g_i}
\end{equation*} 
where $w_{g_i}$ is the weight of goal $g_i$. Assuming that the four goals are equally important, their weights are equal, therefore the weighted sums of the the coalitions are:
\begin{equation*}
\begin{split}
WS(C_0)= 0.25 \times 0.967 + 0.25 \times 0.960 + 0.25 \times 0.975 + 0.25 \times 0.969= 0.96775 \\
WS(C_1)= 0.25 \times 0.968 + 0.25 \times 0.960 + 0.25 \times 0.966 + 0.25 \times 0.971= 0.96625
\end{split}
\end{equation*} 
$C_0$ is preferable as it achieves the system's goals with greater certainty.

Assuming that goals $g_1$ and $g_4$ (delivering the pen to desk $D_a$ and the cutter to desk $D_b$) are more important that $g_2$ and $g_3$ (delivering the piece of paper to desk $D_a$ and the glue to desk $D_b$), and the weights of the four goals are: $w_{g_1}=w_{g_4}=0.4$ and $w_{g_2}=w_{g_3}=0.1$, the weighted sums of the two coalitions are:
\begin{equation*}
\begin{split}
WS(C_0)= 0.4 \times 0.967 + 0.1 \times 0.960 + 0.1 \times 0.975 + 0.4 \times 0.969= 0.9679 \\
WS(C_1)= 0.4 \times 0.968 + 0.1 \times 0.960 + 0.1 \times 0.966 + 0.4 \times 0.971= 0.9682
\end{split}
\end{equation*} 
and in this case $C_1$ is considered as the best coalition.


\section{Related work}\label{sec:related}

This article does not pretend to be the first work bringing together agents and context logics. Two earlier studies (\cite{parsons98,sabater02}) used Multi-Context Systems as a means of specifying and implementing agent architectures. Both studies propose \emph{breaking} the logical description of an agent into a set of contexts, each of which represents a different component of the architecture, and the interactions between these components are
specified by means of bridge rules between the contexts. The first study \cite{parsons98} followed this approach to simplify the construction of a BDI agent, while the second \cite{sabater02} extended it to handle more efficiently implementation issues such as grouping together contexts in modules, and enabling inter-context synchronization. The main difference with our approach is that the focus is not on the internal representation of agents, but rather on their interactions with other agents and the coalitions that they can form.

As we also mention in the Introduction, this study is built on previous work on modeling multiagent systems as MCS and computing coalitions using MCS computational methods and tools, which we first presented in \cite{eumas14}. Here we give more details about the proposed methodology. We also extend it with modeling and reasoning methods from possibilistic reasoning to handle uncertainty in the agents' actions. In another previous study,  we evaluated information exchange in distributed information systems, based on modeling MCS as dependence networks where bridge rules are represented as dependencies \cite{prima13}. Here we do the opposite: we use bridge rules to represent dependencies among agents, and model agents as contexts in MCS.

Several studies from different research areas have focused on the problem of coalition formation including variants of the contract net protocol \cite{gerkey02,lemaire04}, according to which agents break down composite tasks into simpler subtasks and subcontract subtasks to other agents via a bidding mechanism; formal approaches from multiagent systems, e.g. \cite{klusch02,DBLP:journals/ai/ShehoryK98}; solutions from the field of robotics based on schema theory \cite{tang05,zhang13} or synergy \cite{DBLP:journals/ai/LiemhetcharatV14}; approaches inspired from the formation of coalitions in politics \cite{DBLP:conf/aiia/ChellaSRFP03}; and adaptive approaches based on machine learning methodologies \cite{DBLP:conf/atal/AbdallahL04,DBLP:conf/icai/SohL04}. 

Compared to previous approaches, the solution that we propose is the only one that combines the following four characteristics: (a) it allows agents to use different knowledge representation models; (b) based on a non-monotonic reasoning model, it enables representing and reasoning with agents with conflicting goals; (c) by integrating features from possibilistic reasoning, it also allows handling uncertainty in the agents' actions; and (d) it provides both centralized and distributed algorithms for computing coalitions, and can hence be applied in settings with different requirements for information hiding and sharing, related for example to the privacy and security of the agents.

There is also a number of studies that focus on the problem of forming coalitions under uncertainty. There are different kinds of uncertainty that they consider. For example, studies such as \cite{chalkiadakis07,shehory99} deal with uncertainty in the \emph{agent type}, i.e. in the knowledge about the agent's abilities and motivations. Other studies such as \cite{kargin03} focus on the uncertainty in the agents' costs, e.g. in a multi-robot system the relevant costs can be energy, hardware cost or processing time. Finally, the approaches proposed in \cite{soh03,lovekesh06} deal with uncertainty in the value of a coalition, i.e. the costs and payoff of a coalition for each agent taking part in it. Most of the proposed solutions are based on game theory and machine learning algorithms. Our approach, on the other hand, focuses on another kind of uncertainty, i.e. uncertainty in the agents' actions, and our proposed solution combines elements of logic-based and possibilistic reasoning.

\section{Summary and Future Work}\label{sec:summary}

In multiagent systems agents often depend on each other and need to cooperate in order to achieve their goals. In this article, we deal with the problem of computing the alternative coalitions in which the agents may fulfil their goals. Specifically, we propose a representation of multiagent systems based on Multi-Context Systems, according to which agents are modeled as contexts and dependence relations among agents as bridge rules. 
Based on this representation, we compute the equilibria of the MCS, which correspond to the coalitions in which the agents may fulfil their goals. Then, given a set of functional and non-functional requirements such as efficiency, stability and conviviality, we propose ways to select the best coalition. Finally, we extend our model and algorithms with features from possibilistic reasoning, and more specifically possibilistic MCS, in order to model, compute and evaluate the different coalitions taking also into account the uncertainty in the agents' actions. We demonstrate the proposed approach using an example from robotics, in which four different robots need to cooperate in order to perform a given set of tasks. 

In further research, we plan to integrate preferences on agents and goals into our model. Building on previous work on preference-based inconsistency resolution in MCS \cite{bikakisKAIS11,eiterKR10,eiterJELIA10}, we will develop algorithms for preference-based coalition formation in the presence of conflicting goals. We also want to extend our approach with elements of dynamic MCS \cite{dao-tran11}, i.e. schematic bridge rules that are instantiated at run time with concrete contexts. This will enable us to handle changes such as the failure of an agent, the arrival of a new agent or any change in the operating environment. We will also plan to apply and test our methods in different kinds of agent-based systems characterized by the heterogeneity of the participating agents and uncertainty, such as ubiquitous robots and Ambient Intelligence systems. To achieve this we will use existing MCS implementations, such as DMCS \cite{bairakdar10}, a distributed solver for MCS, and MCS-IE \cite{bogl10}, a tool for explaining inconsistencies in MCS, and extend them with features from possibilistic MCS.



\bibliographystyle{spmpsci}      
\bibliography{references}
\end{document}